\documentclass[runningheads]{llncs}

\usepackage{eccv}

\usepackage{eccvabbrv}

\usepackage{graphicx}
\usepackage{booktabs}

\usepackage[accsupp]{axessibility}  %

\newcommand{\algname}{DriveLM-Agent\xspace}
\newcommand{\dataname}{DriveLM\xspace}

\newcommand{\figref}[1]{Fig.~\ref{#1}}
\newcommand{\secref}[1]{Section~\ref{#1}}

\newcommand{\tabref}[1]{Table~\ref{#1}}

\makeatletter
\DeclareRobustCommand\onedot{\futurelet\@let@token\@onedot}
\def\@onedot{\ifx\@let@token.\else.\null\fi\xspace}
\def\eg{e.g\onedot} 
\def\ie{i.e\onedot}

\makeatother

\newcommand{\boldparagraph}[1]{\vspace{0.0cm}\noindent{\bf #1.}}

\definecolor{darkgreen}{rgb}{0,0.7,0}
\definecolor{darkyellow}{rgb}{0.8,0.8,0}
\definecolor{bittersweet}{rgb}{1.0, 0.44, 0.37}
\definecolor{amber}{rgb}{1.0, 0.49, 0.0}
\definecolor{lgray}{rgb}{0.83,0.83,0.83}

\definecolor{color_unlabled}{rgb}{0.0,0.0,0.0}
\definecolor{color_vehicle}{rgb}{0.0,0.0,0.56}
\definecolor{color_road}{rgb}{0.5,0.25,0.5}
\definecolor{color_redlight}{rgb}{1.0,0.0,0.0}
\definecolor{color_person}{rgb}{0.859,0.078,0.234}
\definecolor{color_roadline}{rgb}{0.613,0.914,0.195}
\definecolor{color_sidewalk}{rgb}{0.953,0.137,0.906}
\definecolor{teaser_red}{RGB}{222,112,97}

\definecolor{ellisred}{rgb}{0.87,0.44,0.38} %
\definecolor{ellisgreen}{rgb}{0.69,0.90,0.52} %
\definecolor{elliscyan}{rgb}{0.29,0.77,0.74} %
\definecolor{ellisorange}{rgb}{0.89,0.55,0.28} %
\definecolor{ellisblue}{rgb}{0.41,0.61,0.86} %

\definecolor{tuedgray}{RGB}{56,55,55}
\definecolor{tuelgray}{RGB}{246,246,246}
\definecolor{tuedblue}{RGB}{26,58,91}
\definecolor{tuelblue}{RGB}{133,203,210}
\definecolor{tueoblue}{RGB}{119,221,204}
\definecolor{tueogreen}{RGB}{119,221,159}
\definecolor{tuesgreen}{RGB}{186,213,72}
\definecolor{tueyellow}{RGB}{255,221,0}
\definecolor{tuered}{RGB}{234,75,46}

\newcommand{\tuedblue}[1]{\noindent{\color{tuedblue}{#1}}}
\newcommand{\tuelblue}[1]{\noindent{\color{tuelblue}{#1}}}
\newcommand{\tueoblue}[1]{\noindent{\color{tueoblue}{#1}}}
\newcommand{\tueogreen}[1]{\noindent{\color{tueogreen}{#1}}}
\newcommand{\tuesgreen}[1]{\noindent{\color{tuesgreen}{#1}}}

\newcommand{\cmtt}[1]{{\fontfamily{cmtt}\selectfont #1}}

\mathchardef\mhyphen="2D

\usepackage{hyperref}

\usepackage{orcidlink}

\usepackage{multirow}
\usepackage{multicol}
\usepackage{makecell}
\usepackage{xcolor}
\usepackage{algorithm} 
\usepackage{amssymb}
\usepackage{amsfonts}
\usepackage{graphicx}
\usepackage{booktabs}
\usepackage{xr}
\usepackage{epigraph}
\newcommand\blankfootnote[1]{%
  \let\thefootnote\relax\footnotetext{#1}%
  \let\thefootnote\svthefootnote%
}

\usepackage[T1]{fontenc}
\definecolor{mycitecolor}{rgb}{0, 0.4, 0.7}

\usepackage{comment}
\usepackage{subcaption}
\usepackage{amsmath}
\usepackage{algorithm}
\usepackage{array}
\usepackage{longtable}
\usepackage{xspace}
\usepackage{url}
\usepackage{overpic}
\usepackage{ragged2e}
\usepackage{xpatch}
\usepackage{pifont}
\usepackage{enumitem}
\usepackage{bbm}
\usepackage{colortbl}
\usepackage{soul}
\usepackage{adjustbox}
\usepackage{hhline}
\usepackage{algpseudocode}
\newcommand{\bheading}[1]{{\noindent{\textbf{#1}}}}

\newcommand{\xmark}{\ding{55}}%
\usepackage{stfloats}
\usepackage{xr-hyper}
\usepackage{hyperref}
\usepackage{tcolorbox}
\usepackage{wrapfig}
\tcbuselibrary{breakable}

\usepackage{etoolbox}
\makeatletter
\pretocmd{\chapter}{\addtocontents{toc}{\protect\addvspace{15\p@}}}{}{}
\pretocmd{\section}{\addtocontents{toc}{\protect\addvspace{5\p@}}}{}{}
\pretocmd{\subsection}{\addtocontents{toc}{\protect\addvspace{3\p@}}}{}{}
\makeatother
\usepackage{titletoc}

\begin{document}

\title{\dataname: \\Driving with Graph Visual Question Answering} 

\author{
{Chonghao Sima}$^{4,1\ast}$ \quad
Katrin Renz$^{2,3\ast}$ \quad
Kashyap Chitta$^{2,3}$ \quad
Li Chen$^{4,1}$ \\
Hanxue Zhang$^{1}$ \quad
Chengen Xie$^{1}$ \quad
Jens Beißwenger$^{2,3}$ \quad
Ping Luo$^{4}$ \\
Andreas Geiger$^{2,3\dagger}$ \quad
Hongyang Li$^{1\dagger}$ \\
}

\authorrunning{C.~Sima et al.}
\institute{OpenDriveLab, Shanghai AI Lab, China \and
University of Tübingen, Germany \and Tübingen AI Center, Germany \and University of Hong Kong, China \\
\email{simachonghao@pjlab.org.cn \quad katrin.renz@uni-tuebingen.de}
}

\maketitle
\blankfootnote{$^\ast$Equal contribution. $^\dagger$Project lead.}

\begin{center}
    \centering
    \captionsetup{type=figure}
    \includegraphics[width=\textwidth]{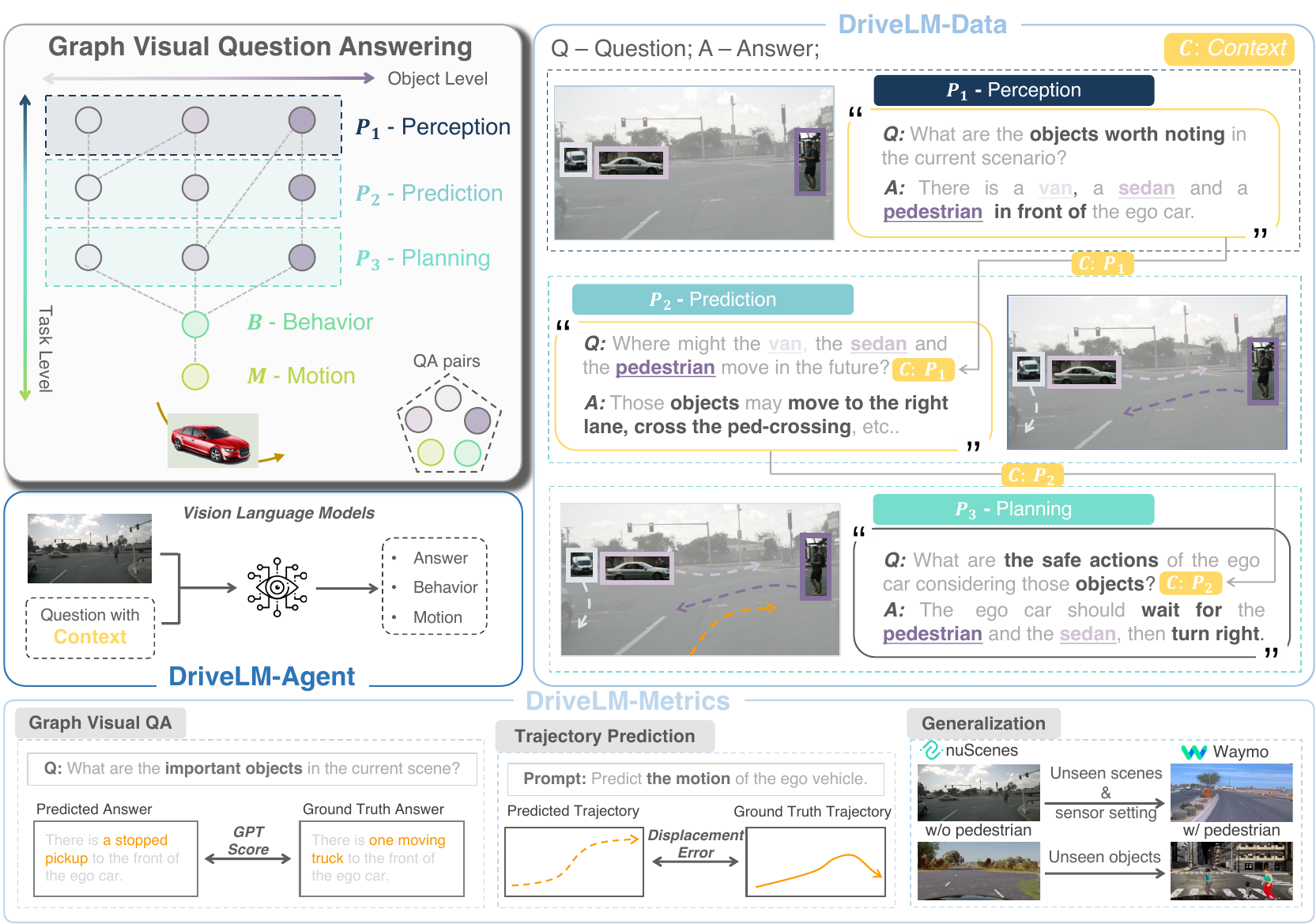}
    \vspace{-15pt}
    \captionof{figure}{\label{fig:teaser}
    We present \textbf{\dataname}:
    A new task, dataset, metrics, and baseline for end-to-end autonomous driving. Inspired by~\cite{chen2023end}, \dataname considers \textbf{{Graph Visual Question Answering (GVQA)}}, where QA pairs are interconnected via logical dependencies at the object-level, \textit{\ie}, interactions between object pairs, and the task-level, \textit{\eg}, \tuedblue{perception} $\rightarrow$ \tuelblue{prediction} $\rightarrow$ \tueoblue{planning} $\rightarrow$ \tueogreen{behavior} (discretized action described in natural language) $\rightarrow$ \tuesgreen{motion} (continuous trajectory). We propose \textbf{\dataname-Data} for training \textbf{\algname}, a baseline for GVQA. We validate its effectiveness using the \textbf{\dataname-Metrics} on challenging settings requiring zero-shot generalization.
    }
    \vspace{-0.2cm}
\end{center}

\begin{abstract}
We study how vision-language models (VLMs) trained on web-scale data can be integrated into end-to-end driving systems to boost generalization and enable interactivity with human users. While recent approaches adapt VLMs to driving via single-round visual question answering (VQA), human drivers reason about decisions in multiple steps. Starting from the localization of key objects, humans estimate object interactions before taking actions. The key insight is that with our proposed task, Graph VQA, where we model graph-structured reasoning through perception, prediction and planning question-answer pairs, we obtain a suitable proxy task to mimic the human reasoning process. We instantiate datasets (DriveLM-Data) built upon nuScenes and CARLA, and propose a VLM-based baseline approach (DriveLM-Agent) for jointly performing Graph VQA and end-to-end driving. The experiments demonstrate that Graph VQA provides a simple, principled framework for reasoning about a driving scene, and DriveLM-Data provides a challenging benchmark for this task. Our DriveLM-Agent baseline performs end-to-end autonomous driving competitively in comparison to state-of-the-art driving-specific architectures. Notably, its benefits are pronounced when it is evaluated zero-shot on unseen sensor configurations. Our question-wise ablation study shows that the performance gain comes from the rich annotation of prediction and planning QA pairs in the graph structure. All data, models and an official evaluation server are available at \url{https://github.com/OpenDriveLab/DriveLM}.
\end{abstract}
\section{Introduction}
\label{sec:intro}
\vspace{-4pt}
Current Autonomous Driving (AD) stacks are still lacking crucial capabilities~\cite{chen2023end,chib2023recent}. One key requirement is generalization, which involves the ability to handle unseen scenarios or unfamiliar {sensor configurations}. A secondary requirement pertains to the interaction of these models with humans, highlighted for example by EU regulations that mandate explainability in deployment~\cite{atakishiyev2023explaining}. Furthermore, unlike today's AD models, humans do not navigate based on geometrically precise bird's-eye view (BEV) representations~\cite{hu2023_uniad,Chitta2023PAMI,li2022bevsurvey}. Instead, humans implicitly perform object-centric perception, prediction, and planning (which we refer to as $P_{1-3}$): a rough identification and localization of key objects, followed by reasoning about their possible movement and aggregation of this information into a driving action~\cite{Marr82,spelke2007core}. 

Simultaneously, another field has been forging ahead: Vision-Language Models (VLMs)~\cite{li2023blip2,zhang2023llamaadapter,liu2023llava,wang2023visionllm}. These models have several strengths. First, they hold a base understanding of the world from internet-scale data that could potentially facilitate generalization for planning in AD. In fact, this sort of generalization has already been achieved by VLMs for simpler robotics tasks~\cite{zitkovich2023rt,driess2023palme}. Second, the use of language representations as an input and output offers a platform for human-friendly interaction with these models, unlike bounding boxes or trajectories that are more common to current methods~\cite{li2022bevformer,hu2022stp3,Renz2022CORL,Dauner2023CORL}. Finally, VLMs are able to make decisions in multiple steps linked by logical reasoning~\cite{zitkovich2023rt,wei2022chainofthought,wang2023selfconsistency,yao2023tree,besta2023got,ding2023everything}. Importantly, even though they reason in multiple separate steps, VLMs are end-to-end differentiable architectures, a characteristic that is highly desirable for autonomous driving~\cite{chen2023end}.

Recent work towards enabling the application of VLMs to AD systems falls into two categories: scene-level or single object-level Visual Question Answering (VQA). Scene-level VQA refers to the task of describing the driving behavior by one or two supporting reasons, \textit{\eg}, {``The car is moving into the right lane because it is safe to do so.''}~\cite{kim2018bddx,kim2019had}. Single object-level VQA formulates the understanding of the ego vehicle's response to a single object by a chain of QAs in the form of ``what-which-where-how-why'', \textit{\eg}, {``The ego vehicle stops because there is a pedestrian in a white shirt crossing the intersection in front of the ego vehicle and it does not want to crash into the pedestrian.''}~\cite{sachdeva2023rank2tell,malla2023drama,qian2023nuscenesqa}. Unfortunately, neither of these paradigms provides a suitable proxy task to mimic the $P_{1-3}$ reasoning process in humans, who consider multiple objects and reason about each in multiple steps. Therefore, in this paper, we propose a new task, along with corresponding datasets and a baseline model architecture (\figref{fig:teaser}).

\smallskip
\boldparagraph{Task} \textbf{{Graph Visual Question Answering (GVQA)}} involves formulating $P_{1-3}$ reasoning as a series of question-answer pairs (QAs) in a directed graph. Its key difference to the aforementioned VQA tasks for AD is the availability of logical dependencies between QAs which can be used to guide the answering process. GVQA also encompasses questions regarding {behavior} and {motion} planning, with dedicated {{metrics}} (details in \secref{sec:data}).

\begin{figure*}[t!]
  \centering
  \includegraphics[width=\linewidth]{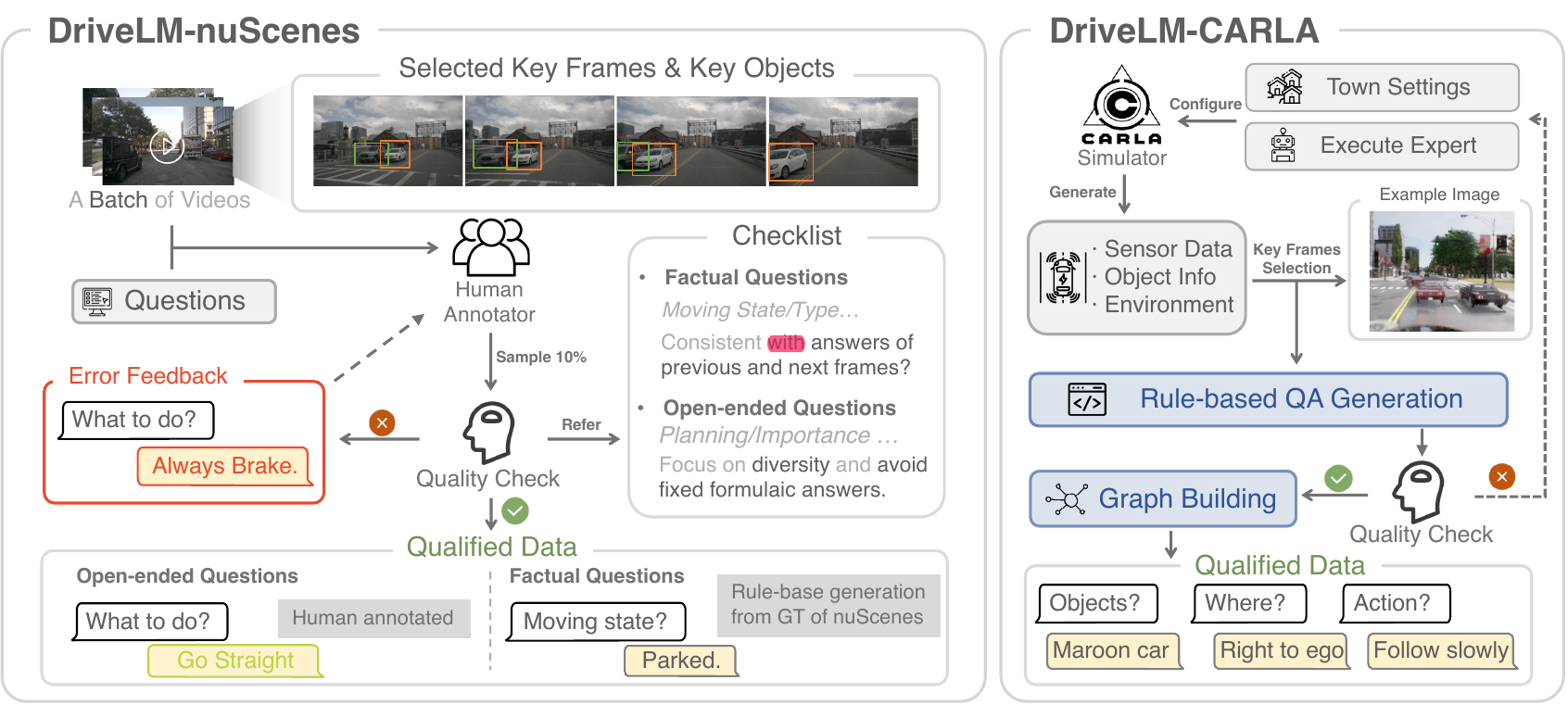}
  \vspace{-15pt}
  \caption{\textbf{Annotation Pipeline:} 
  In \dataname-nuScenes, we adopt a semi-rule-based QA labeling pipeline, where both the ground truth annotation in nuScenes/OpenLane-V2 and feedback from human annotators are used. A critical part of our pipeline is the multi-round quality check, which guarantees high data quality at reasonable costs. In \dataname-CARLA, we meet the same standards while exploiting a fully rule-based QA labeling pipeline instead, using a new expert algorithm called PDM-Lite.
  }
  \label{fig:data_collection}
\vspace{-0.4cm}
\end{figure*}

\smallskip
\boldparagraph{Datasets} \textbf{{\dataname-nuScenes}} and \textbf{{\dataname-CARLA}} consist of annotated QAs, arranged in a graph, linking images with driving behavior through logical reasoning. In comparison to existing benchmarks, they provide significantly more text annotations per frame (\tabref{tab:dataset} and \figref{fig:data_collection}). {Furthermore, as an integral component of \dataname-CARLA, we build PDM-Lite \cite{Beißwenger2024PdmLite}, the first working rule-based expert algorithm for CARLA Leaderboard 2.0.} We pair these training datasets with challenging test data for evaluating zero-shot generalization. 

\smallskip
\boldparagraph{Baseline} \textbf{{\algname}} employs a trajectory tokenizer that can be applied to any general %
VLM~\cite{li2023blip2,liu2023llava,zhang2023llamaadapter,peng2023kosmos}, coupled with a graph prompting scheme that models logical dependencies as context inputs for VLMs. The result is a simple methodology to effectively repurpose VLMs for end-to-end AD (\secref{sec:method}).
Our experiments provide encouraging results. We find that GVQA on \dataname is a challenging task, where current methods obtain moderate scores and better modeling of logical dependencies is likely necessary to achieve strong QA performance. Even so, \algname already performs competitively to state-of-the-art driving-specific models~\cite{hu2023_uniad} when tested in the open-loop planning setting, despite its task-agnostic architecture. Furthermore, employing a graph structure improves zero-shot generalization, enabling \algname to correctly handle unseen deployment on Waymo data~\cite{sun2019waymo} after training only on nuScenes~\cite{caesar2019nuscenes} data. {Finally, with a question-wise analysis we find that the QA pairs from the prediction and planning stages help the final driving decision most}. From these results, we believe that improving GVQA holds great potential towards building autonomous driving agents with strong generalization.

\section{\dataname: Task, Data, Metrics}
\label{sec:data}
\vspace{-4pt}

Human drivers usually decompose their decision-making process into distinct stages that follow a logical progression which encompasses the identification and localization of key objects, their possible future action and interaction, and ego planning based on all this information~\cite{macadam2003understanding,groeger2013understanding}. This inspires us to propose the GVQA as the critical ingredient of \dataname, which serves as a suitable proxy task to mimic the human reasoning process. Within this section, we illustrate the formulation of the {GVQA task} (\secref{sec:graph}), introduce {\dataname-Data} (\secref{sec:fact}) to exemplify the instantiation of GVQA using prominent driving datasets, and overview the {\dataname-Metrics} used for evaluation (\secref{sec:task}). 
{To encourage further research in this direction, an official evaluation server (with a public leaderboard) will be set up to benchmark different methods and discover more insights about combining language models with autonomous driving.}

\begin{table*}[t]
\caption{\textbf{Comparison of \dataname-nuScenes \& -CARLA with Existing Datasets.} \dataname-Data significant advances annotation quantity, comprehensiveness (covering \textbf{perception, prediction and planning}), and logic (chain to \textbf{graph}). $\dag$ full dataset, $\ddag$ keyframe dataset, $^{*}$ semi-rule-based labeling (w/ human annotators), $^{**}$ fully-rule-based (no human annotators). - means publicly unavailable.
} 
\vspace{-10pt}
\centering
\small
\resizebox{\textwidth}{!}{
    \setlength\tabcolsep{1pt}
    \begin{tabular}{l||ccc|ccc|c}
        \toprule 
        & & & & & & & \\
        \multirow{-2}{*}{\textbf{Dataset}} & \multirow{-2}{*}{\shortstack{ \textbf{Source}\\ \textbf{Dataset}} } & \multirow{-2}{*}{\textbf{\# Frames}} &  \multirow{-2}{*}{\shortstack{ \textbf{Avg. QA}\\ \textbf{per Frame}} } &   \multirow{-2}{*}{ \textbf{Perception}}  & \multirow{-2}{*}{ \textbf{Prediction} }     & \multirow{-2}{*}{\textbf{Planning}}    & 
        \multirow{-2}{*}{\textbf{Logic}}  \\
        \midrule
        nuScenes-QA~\cite{qian2023nuscenesqa}  & nuScenes & 34,149 &13.5 & 460k$^{**}$  & \xmark  & \xmark   & None  \\
        nuPrompt~\cite{wu2023nuprompt} & nuScenes & 34,149 & 1.0 & 35k$^{*}$ &  \xmark   & \xmark     &  None   \\ 
        HAD~\cite{kim2019had} & HDD & 25,549 & 1.8 & 25k & \xmark  & 20k  &  None   \\ 
        BDD-X~\cite{kim2018bddx}  & BDD & 26,228 & 1 & 26k  & \xmark & \xmark  &  None \\ 
        LingoQA~\cite{marcu2023lingoqa} & LingoQA & 28,000 & 15.3  & - & - & - & None \\
        DRAMA~\cite{malla2023drama}             & DRAMA & 17,785 & 5.8 & 85k  &  \xmark  & 17k  & Chain \\
        Rank2Tell~\cite{sachdeva2023rank2tell} & Rank2Tell & 5,800 & - & - & \xmark &  -  & Chain  \\

        \midrule
        \rowcolor[gray]{0.95}{{\dataname-nuScenes}} & nuScenes & 4,871 & \textbf{91.4} & 144k$^{*}$ & 153k &  146k  & \textbf{Graph}   \\
        \rowcolor[gray]{0.95}{{\dataname-CARLA\dag}} & CARLA & {64,285} & {24.4} & {697k}$^{**}$ & {311k}$^{**}$ &  {558k}$^{**}$  & \textbf{Graph}   \\
         \rowcolor[gray]{0.95}{{\dataname-CARLA\ddag}} & CARLA & {5,721} & {24.8} & {63k}$^{**}$ &  {28k}$^{**}$  & {51k}$^{**}$ & \textbf{Graph}   \\
        \bottomrule
\end{tabular} }
\label{tab:dataset}
\vspace{-0.5cm}
\end{table*}

\subsection{\dataname-Task: GVQA}
\label{sec:graph}

We organize all the question-answer pairs (QAs) for an image frame into a graph. {We use the terminology ``graph-structured'' to refer to a directed acyclic graph (DAG), e.g., the current question can get context from (multiple) parent and grandparent nodes.
The graph $G\!=\!(V, E)$ contains a set of vertices $V$}, where each vertex represents a QA pair $v\!=\!(q,a)$ associated with one or more key objects in the scenario. The key difference between GVQA and ordinary VQA is that the QAs in GVQA have logical dependencies, which we formulate as the edges between the vertices. $E\subseteq\!V\!\times\!V$, is a set of directed edges, where each edge $e\!=\!(v_p, v_c)$ connects the parent QA and the child QA. We formulate the edge set $E$ by incorporating two dimensions: object-level and task-level edges. At the object level, we construct the logical edges $e\!\in\!E$ to represent the impact of interactions between different objects. For example, the planning QA node for the sedan is influenced by the perception QA node of the pedestrian in the illustration from \figref{fig:teaser} (center). At the task-level, we establish the logical edges $e\!\in\!E$ to capture the logical chain of different reasoning stages:
\begin{itemize}
    \item \textbf{Perception} ($P_1$): identification, description, and localization of key objects in the current scene.
    \item \textbf{Prediction} ($P_2$): estimation of possible action/interaction of key objects based on perception results.
    \item \textbf{Planning} ($P_3$): possible safe actions of the ego vehicle.
    \item \textbf{Behavior} ($B$): classification of driving decision.
    \item \textbf{Motion} ($M$): waypoints of ego vehicle future trajectory.
\end{itemize}

\smallskip
The concepts of perception, prediction, and planning ($P_{1-3}$) are similar to those in end-to-end AD~\cite{chen2023end}, while the concepts of motion and behavior are based on the ego vehicle future trajectory. Specifically, we define the motion $M$ as the ego vehicle future trajectory, which is a set of $N$ points with coordinates $(x,y)$ in bird's-eye view (BEV), denoted as $M$ = $\{(x_0,y_0), (x_1,y_1) , ..., (x_N, y_N)\}$. Each point is the offset between the future position and the current position by a fixed time interval. Then, the distance for $x,y$ at each time interval is computed as: 
\begin{equation}
    {\{x,y\}}_{\text{dist}} = \{ (\delta_{x,1},\delta_{y,1}),...,(\delta_{x,N},\delta_{y,N}) \},
    \label{eq:xy_dist}
\end{equation}
where $\delta_{x,i}\!=\!x_i\!-\!x_{i-1}$ and $\delta_{y,i}\!=\!y_i\!-\!y_{i-1}$, for $i\!=\!1,2,\ldots,N.$ The goal of the behavior representation is to serve as an interface from $P_{1-3}$ to $M$. To obtain a behavior representation, we map the mean of $x_{\text{dist}}$ and $y_{\text{dist}}$ to one of the predefined bins, where each bin corresponds to a category in either speed or steering. These are denoted as $B_{sp}$ and $B_{st}$ respectively. In this work, we consider 5 bins: 
\vspace{-5pt}
\begin{align*}
    B_{sp} &\in \{ \texttt{fast}_2, \texttt{fast}_1, \texttt{moderate}, \texttt{slow}_1, \texttt{slow}_2 \}, \\
    B_{st} &\in \{ \texttt{left}_2, \texttt{left}_1, \texttt{straight}, \texttt{right}_1, \texttt{right}_2 \},
\end{align*}
where the number in the subscript indicates the intensity. The combination of the speed and steering categories for a trajectory form its {behavior} category as $B\!=\!(B_{sp}, B_{st})$. While we use a simple definition of $B$ as a starting point for research on driving with VLMs, we note that our formulation supports the incorporation of more abstract behaviors such as a lane changes or overtaking.

\subsection{\dataname-Data}
\label{sec:fact}

In order to provide comprehensive and accurate QAs with the graph structure defined in~\secref{sec:graph}, we introduce {\dataname-nuScenes} and {\dataname-CARLA}. Since there are significant disparities between nuScenes and CARLA, the collection methods and statistics of these datasets differ.

\smallskip
\noindent\textbf{\dataname-nuScenes.} We divide the annotation process into three steps: selecting key frames from video clips, choosing key objects within these key frames, and subsequently annotating the frame-level $P_{1-3}$ QAs for these key objects. A portion of the Perception QAs are generated from the nuScenes~\cite{caesar2019nuscenes} and OpenLane-V2~\cite{wang2023openlanev2} ground truth, while the remaining QAs are manually annotated.
{The question templates for the manual annotations were designed by 5 domain experts accounting for how humans make driving decisions. Annotators are prompted with all question templates for each frame, encouraged to answer all questions, but provided a {skip option} to account for possible incompatibility.}
As we manually annotate the vast majority of data in \dataname-nuScenes, quality is particularly crucial for this portion. When annotating, we conduct multiple rounds of rigorous quality checks. In each round, we categorize the data into different batches and inspect ten percent of the data in each batch. If the qualification rate of manually annotated data in this ten percent does not meet expectations, we request the annotators to re-label all data in the batch. In~\figref{fig:data_collection}, we showcase an example of the QA annotation pipeline, where all questions undergo quality checks according to our standards. As a result, \dataname-nuScenes stands out from previously proposed datasets with its larger scale, greater comprehensiveness, and more complex structure (See~\tabref{tab:dataset}). These QAs cover various aspects of the driving process, ranging from perception and prediction to planning, providing a comprehensive understanding of autonomous driving scenarios (details in the supplementary material). 

\smallskip
\noindent\textbf{\dataname-CARLA Expert.} We collect data using CARLA 0.9.14 in the Leaderboard 2.0 framework~\cite{Dosovitskiy17}. {Leaderboard 2.0 contains a large set of new driving scenarios compared to its predecessor, Leaderboard 1.0. However, as of now, there is no existing method to collect training data at scale in Leaderboard 2.0. For example, the privileged rule-based expert used by TransFuser++~\cite{Jaeger2023ICCV}, a state-of-the-art method in Leaderboard 1.0, obtains a Driving Score (DS) of only 2\% on the 8+ kilometer long official validation routes of Leaderboard 2.0. We build a new expert algorithm, PDM-Lite, that can handle the new challenges in Leaderboard 2.0. Our approach is similar to PDM-Closed~\cite{Dauner2023CORL}, a rule-based planner for nuPlan~\cite{Caesar2021CVPRW}. PDM-Lite uses the Intelligent Driver Model (IDM)~\cite{treiber2000idm} to obtain a target speed based on the leading vehicle, pedestrian, stop sign or traffic light. Unlike the 16 proposals from which one is selected via a complex cost function in PDM-Closed, we use only two proposals and a simpler cost function based on the TransFuser++ expert~\cite{Jaeger2023ICCV}, giving a light-weight planner suitable for scalable QA generation. PDM-Lite obtains an improved DS of 44\% on the official CARLA validation routes. More details can be found in the supplementary.}

\smallskip
\noindent\textbf{\dataname-CARLA QA Generation.} For collecting the \dataname-CARLA dataset, we set up a series of routes in urban, residential, and rural areas and execute PDM-Lite on these routes. During this process, we collect the necessary sensor data, sample keyframes, generate relevant QAs based on privileged information about objects and the scene, and organize the logical relationships to connect this series of QAs into a graph. We generate data and labels at {4} FPS {and extract keyframes based on changes in the decision of the expert (e.g. when the expert changes from acceleration to braking).} The rule-based annotation pipeline is illustrated in Fig.~\ref{fig:data_collection}. {During data collection, we extract privileged information from the simulator about the status of the static and dynamic objects in the scene as well as the triggered rules of the expert. We use all 38 scenarios except for InterurbanAdvancedActorFlow, MergerIntoSlowTraffic, and VehicleTurningRoute to create questions and answers with hand-crafted sentence templates based on the information we extract from the simulator.} The exact questions with their graph structure can be found in the supplementary. Our annotation process has the advantage of straightforward scalability since we only need to define route and scenario settings in CARLA and the subsequent steps can be executed automatically. 
Including {1.6M QAs (with a straightforward scaling recipe)}, our \dataname-CARLA stands out as the largest driving-language benchmark in terms of total textual content among existing benchmarks as shown in Table~\ref{tab:dataset}.

\vspace{-0.2cm}

\subsection{DriveLM-Metrics}
\label{sec:task}

To evaluate GVQA, the \dataname-Metrics consist of three components for evaluating motion $M$, behavior $B$, and  $P_{1-3}$. For measuring the performance of the {motion} stage, we use standard metrics from the nuScenes and Waymo benchmarks: average and final displacement error, (\textbf{ADE, FDE}), and the \textbf{collision rate} on the predicted trajectory, following UniAD~\cite{hu2023_uniad}. We evaluate {behavior} predictions by the \textbf{classification accuracy}, along with a breakdown of the overall accuracy into its steering and speed components. Finally, we measure the $P_{1-3}$performance using two metrics. \textbf{SPICE}~\cite{anderson2016spice} is a prevailing metric used in VQA and image captioning, which calculates the structure similarity of predicted texts with ground truth while ignoring the semantic meanings. Simultaneously, we employ \textbf{GPT Score} to measure the semantic alignment of answers and complement the SPICE metric. Specifically, the question, the ground truth answer, the predicted answer, and a prompt asking for a numerical score of the answer are sent to ChatGPT-3.5~\cite{ouyang2022training, chatgpt}. We parse the text returned to get the score, where a higher score indicates better semantic accuracy.

\section{\algname: A GVQA Baseline}
\label{sec:method}
\vspace{-4pt}

In this section, we present \algname, a baseline approach for the GVQA task detailed in \secref{sec:data}. \algname is built upon a general vision-language model and can therefore exploit underlying knowledge gained during pre-training. Our overall goal involves translating an image into the desired ego vehicle motion ($M$) through the different stages of VQA ($P_1, P_2, P_3, B$). For this, we choose BLIP-2~\cite{li2023blip2} as our base VLM due to its simplicity in architecture and flexibility in fine-tuning, but our approach can be applied agnostically to other VLMs.

\begin{figure}[t!]
    \centering
    \begin{minipage}[c]{0.55\textwidth}
      \includegraphics[width=\textwidth]{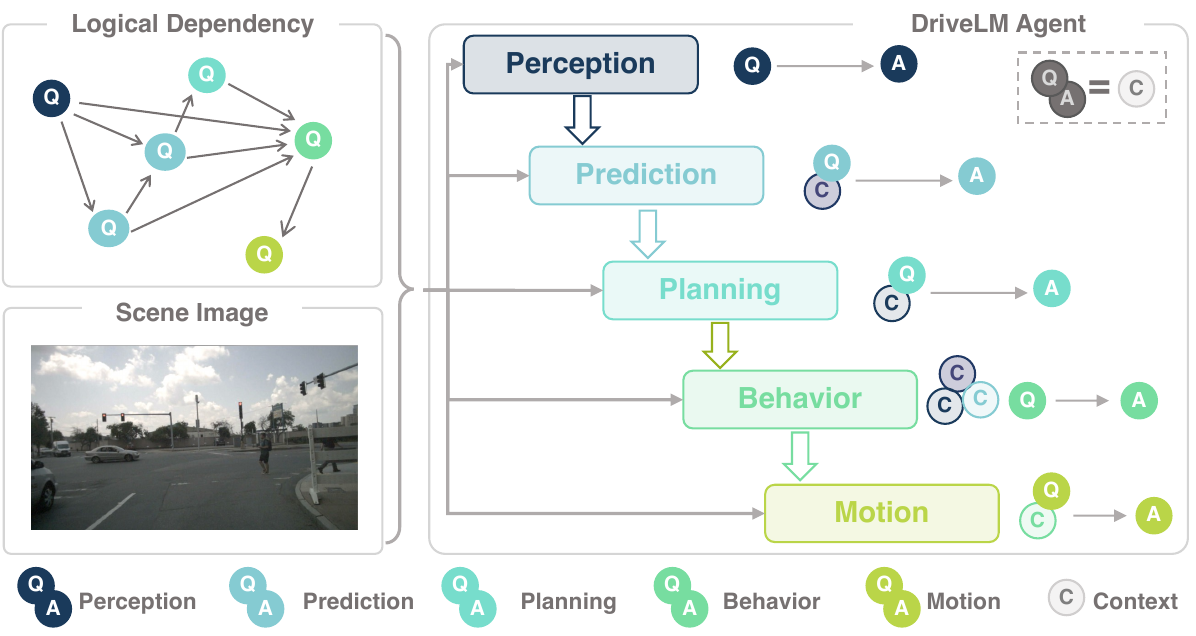}
    \end{minipage} ~ 
    \begin{minipage}[c]{0.33\textwidth}
      \caption{\textbf{\algname Pipeline.} Given the scene image, a VLM performs prompting with context to model the logical dependency among the five QA stages. Context is built using preceding QAs, and can have one or more sources.
  }
    \label{fig:model_pipeline}
    \end{minipage}
    \vspace{-.4cm}
\end{figure}

As shown in \figref{fig:model_pipeline}, \algname can be decomposed into several stages: (1) $P_{1-3}$, \textit{\ie}, \textit{perception, prediction, planning}, serve as the foundational layers to understand the scene and reason about its structure. (2) The \textit{behavior} stage aggregates crucial information from the $P_{1-3}$ into a description of the desired driving action in language space. (3) Finally, the \textit{motion} stage translates the behavior into an executable driving trajectory. To implement the logical dependency between each linked QA, we propose to use context between connected nodes in the GVQA graph. We expand on this idea in the following.

\vspace{-.2cm}
\subsection{Prompting with Context}
\label{sec:context}

Directly translating images to motion as in~\cite{Prakash2021CVPR,Chitta2021ICCV} is extremely challenging. Motivated by the tendency of humans to perform a multi-step reasoning process, we propose to use a similar strategy for VLM-based driving. By doing so, we facilitate the retrieval of knowledge stored in LLMs and improve explainability.

More precisely, the model is designed to use answers from the previous steps in the reasoning process as the {context} for the following questions. For each edge $e=\!(v_p, v_c)\!\in\!E$, we append the QA from the parent node $v_p$ to the question of the current node $v_c$ with a prefix ``\textit{Context: }''. The context can also contain QAs from multiple preceding nodes in which case we concatenate all QAs to one context sequence. It is worth noting that the context is only one possible implementation to formulate logical dependency in GVQA, which we select due to its simplicity. With this scheme, we pass forward relevant information based on the logical dependencies established by the graph.

Note that the size and structure of the graph during inference is a design choice of the algorithm, which can be adapted based on the task or available compute budget. We use this property to train on all available QAs, but perform inference on specific subgraphs, where the questions are sampled using heuristics. For more details, please refer to the supplementary material.

\smallskip
\boldparagraph{Context Aggregation through Behavior} 
Driving encompasses a wide array of potential situations that require an appropriate response. However, despite the diversity of these circumstances, it is interesting to note that almost all events involve decisions that can be discretized into a set of behaviors. For example, applying the brakes appropriately may address various situations such as a red light signal, a stop sign, or the presence of an object ahead of the vehicle. The focus of our behavior stage is to generate such a behavior: a statement in natural language that articulates the vehicle's intended movement. 
As described in~\secref{sec:graph}, this description effectively serves as a reflective step wherein the model summarizes all crucial information from the graph. Thus, we propose to use all possible sources of context for predicting behavior, \textit{\ie}, all the QAs in $P_{1-3}$. 
We empirically observe that this design is crucial for driving with VLMs. 

\vspace{-.2cm}
\subsection{Trajectory Tokenization for Motion}
\label{sec:traj_token}

Since it is non-trivial to output fine-grained numerical results using general VLMs,  RT-2~\cite{zitkovich2023rt} handles robotic actions based on a specialized trajectory tokenization module. We use this approach to enable \algname to take as input the image and behavior description and output trajectories. Specifically, we divide the coordinates of waypoints into 256 bins empirically based on statistics of train set trajectories. We re-define tokens in the BLIP-2 language tokenizer, establishing tokens for each bin, and fine-tune the VLM on the redefined vocabulary. For simplicity, we use the same VLM architecture (BLIP-2) to perform this task, but with independent LoRA weights and trained on a dataset consisting of only the QAs for this motion stage. Thus, it is possible to perform this functionality using a lightweight LLM~\cite{radford2019language} or driving-specific architecture that accepts a command as an input~\cite{hu2022stp3, wu2022trajectoryguided}.

\section{Experiments}
\label{sec:experiments}
\vspace{-4pt}

In this section, we present our experimental results that aim to address the following research questions: (1) How can VLMs be effectively repurposed for end-to-end autonomous driving? (2) Can VLMs for driving generalize when evaluated with unseen sensor setups? {(3) What is the effect of each question in perception, prediction and planning on the final behavior decision?} (4) How well do VLMs perform perception, prediction, and planning via GVQA? 
{Experiments about more VLMs on \dataname-nuScenes and generalization to unseen objects on \dataname-CARLA are included in the supplementary material.}

\smallskip
\boldparagraph{Setup} We now briefly overview the key implementation details for the two settings used in our experiments (additional details are provided in the supplementary material). All fine-tuning is implemented with LoRA~\cite{hu2021lora}. On \dataname-nuScenes, we finetune BLIP-2 on the \texttt{train} split for 10 epochs. We use a batch size of 2 for each GPU, and the entire training process spans approximately 7 hours with 8 V100 GPUs. We train BLIP-2 on {the keyframes of the} \texttt{train} split of \dataname-CARLA for 6 epochs. This takes 6 hours on 4 A100 GPUs.

\subsection{VLMs for End-to-End Driving}
\label{sec:exp_arch}

In our first experiment, we aim to assess the ability of VLMs to perform open-loop planning on \dataname-nuScenes. In particular, we investigate the impact of the context provided to the behavior and motion stages. Given sensor data (and in the case of VLM methods, a text input), the model is required to predict the ego-vehicle future trajectory in the form of waypoints. {Though open-loop planning suffers from a mismatched data distribution~\cite{li2023ego}, we try to alleviate this by evaluating on only the key frames annotated in \dataname-nuScenes.}

\smallskip
\boldparagraph{Baselines} As a reference for the task's difficulty, we provide a simple \textbf{Command Mean} baseline. Each frame in nuScenes is associated with one of 3 commands, `turn left', `turn right', or `go straight'. We output the mean of all trajectories in the training set whose command matches the current test frame command. Further, we compare our approach to the current state-of-the-art on nuScenes, UniAD~\cite{hu2023_uniad}. Besides its original setting which requires video inputs, we train a single-frame version (`\textbf{UniAD-Single}') for a fair comparison. Finally, \textbf{BLIP-RT-2} denotes BLIP-2~\cite{li2023blip2} fine-tuned on \dataname-Data with the trajectory tokenization scheme described in \secref{sec:traj_token} for only the motion task. This acts as an indicator for the performance when using an identical network architecture as \algname, but no context inputs or VQA training data.

\smallskip
\boldparagraph{\algname} We consider 3 variants of \algname incorporating our proposed changes in steps: (1) a 2-stage version that predicts behavior and then motion (as described in \secref{sec:graph}), but without any $P_{1-3}$ context for behavior prediction (`None'); (2) a `Chain' version that builds the $P_{1-3}$ graph, but only passes the final node ($P_3$) to the behavior stage; (3) the full model (`Graph') that uses all QAs from $P_{1-3}$ as context for $B$.

\begin{table}[t!]
\caption{\textbf{Open-loop Planning on \dataname-nuScenes and Zero-shot Generalization across Sensor Configurations on Waymo.} Under nuScenes, using {Behavior ($B$) as context for Motion ($M$) enables end-to-end driving with VLMs on par with UniAD-Single, a state-of-the-art driving-specific architecture. Under Waymo, we randomly sampled 1k frames from the Waymo \texttt{val} set after training on \dataname-nuScenes. \algname outperforms UniAD-Single and benefits from graph context.
}}
\vspace{-8pt}
\centering
\resizebox{\textwidth}{!}{
\begin{tabular}{l|cc|ccc|cc|ccc|cc}
\toprule
    \multirow{3}{*}{\textbf{Method}} & {\textbf{Behavior}} & {\textbf{Motion}} & \multicolumn{5}{c|}{\textbf{nuScenes}} & \multicolumn{5}{c}{\textbf{Waymo}} \\
    \hhline{~~~----------}
    & {\textbf{($B$)}} & {\textbf{($M$)}} & \multicolumn{3}{c|}{\textbf{Behavior ($B$)}} & \multicolumn{2}{c|}{\textbf{Motion ($M$)}} & \multicolumn{3}{c|}{\textbf{Behavior ($B$)}} & \multicolumn{2}{c}{\textbf{Motion ($M$)}} \\
    & \textbf{Context} & \textbf{Context} & Acc. $\uparrow$ & Speed $\uparrow$  & Steer $\uparrow$ & ADE $\downarrow$ & Col. $\downarrow$ & Acc. $\uparrow$ & Speed $\uparrow$  & Steer $\uparrow$ & ADE $\downarrow$ & Col. $\downarrow$  \\
    \midrule
    Command Mean & - & - & - & - & - & 4.57 & 5.72 & - & - & - & 7.98 & 11.41 \\
    UniAD-Single & - & - & - & - & - & 1.80 & 2.62 & - & - & - & 4.16 & 9.31\\
    BLIP-RT-2 & - & - & - & - & - & 2.63 & 2.77 & - & - & - & 2.78 & 6.47 \\
    \midrule
    \rowcolor[gray]{0.9} & None & $B$ & \textbf{61.45} & \textbf{72.20} & \textbf{84.73} & \textbf{1.39} & \textbf{1.67} &  35.70 & 43.90 & 65.20 & 2.76 & 6.59 \\
    \rowcolor[gray]{0.9} \algname & Chain & $B$ & 50.43 & 60.32 & 75.34 & 2.07 & 2.08 & 34.62 & 41.28 & 64.55 & 2.85 & 6.89\\
    \rowcolor[gray]{0.9} & Graph & $B$ & 57.49 & 69.89 & 80.63 & 1.74 & 1.89 & \textbf{39.73} & \textbf{54.29} & \textbf{70.35} & \textbf{2.63} & \textbf{6.17}\\
    \midrule
    \textit{UniAD~\cite{hu2023_uniad}} & \textit{-} & \textit{-} & \textit{-} & \textit{-} & \textit{-} & \textit{0.80} & \textit{0.17} & \textit{-} & \textit{-} & \textit{-} & \textit{-} & \textit{-}\\
\bottomrule
\end{tabular}}
\label{tab:arch}
\vspace{-0.4cm}
\end{table}

\smallskip
\boldparagraph{Results} We show results for the above methods in \tabref{tab:arch}. Among the baselines, BLIP-RT-2 is unable to match UniAD-Single (though both perform well relative to Command Mean). This shows that the single-stage approach without any reasoning is unable to compete with the prior state-of-the-art on nuScenes. However, the proposed \algname, which predicts behavior as an intermediate step for motion, provides a significant boost in performance, surpassing UniAD-Single. This indicates that with the appropriate prompting, VLMs can be surprisingly competitive for end-to-end driving. Interestingly, in the experimental settings that do not involve generalization, the Chain and Graph versions of \algname do not provide any further advantage over no context. Further, single-frame VLMs fall short compared to the privileged video-based UniAD model, indicating that VLMs with video inputs may be necessary for this task. We provide video-input VLM result in the supplementary material.

\begin{figure*}[t]
    \centering
    \includegraphics[width=\textwidth]{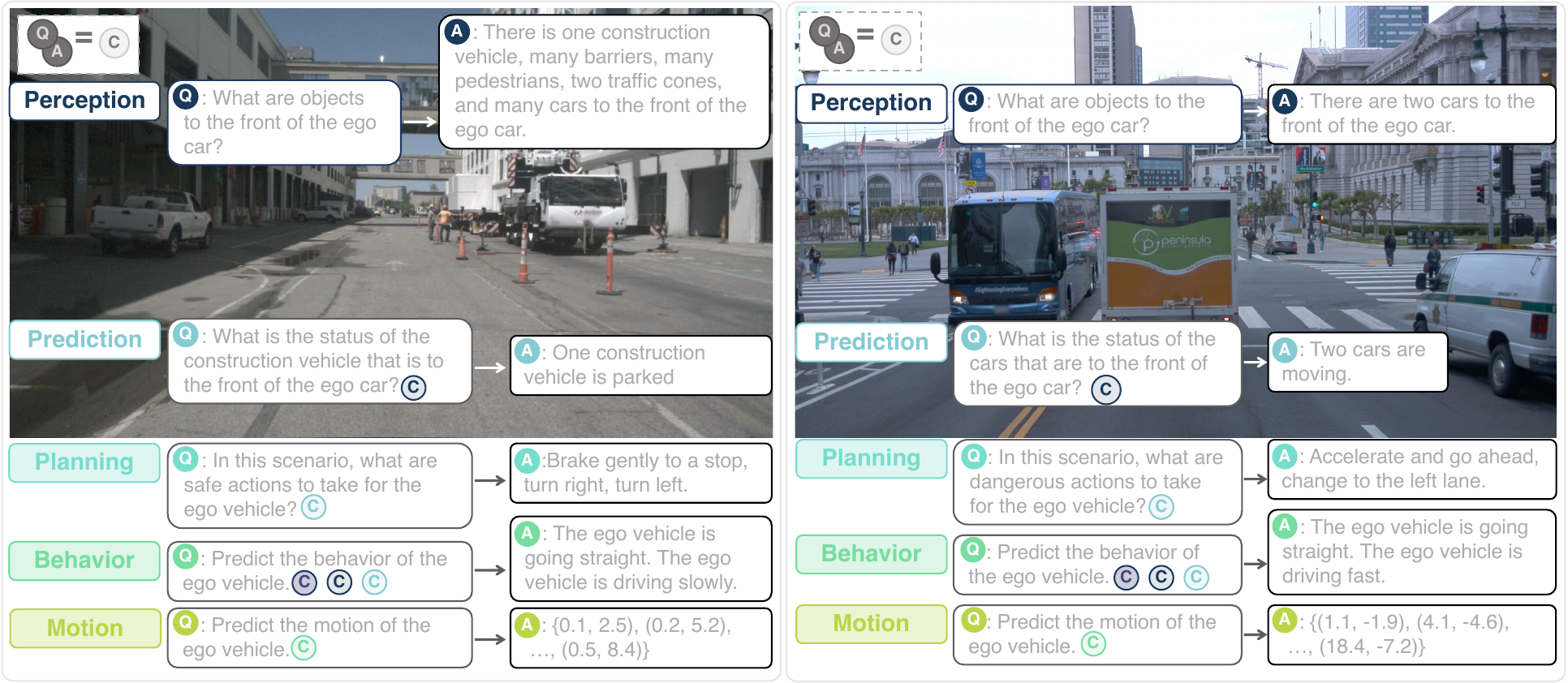}
    \vspace{-15pt}
    \caption{\textbf{Qualitative Results of \algname.} (Left) \dataname-nuScenes \texttt{val} frame, (Right) Waymo \texttt{val} frame. We show the questions (Q), context (C), and predicted answers (A). 
    \algname's outputs are easy to interpret for human users. 
    }
    \label{fig:qualitative}
    \vspace{-0.4cm}
\end{figure*}

\subsection{Generalization Across Sensor Configurations}
\label{sec:exp_waymo}

As a more challenging setting for evaluating the models from \secref{sec:exp_arch}, we now apply them without further training to a new domain: the Waymo dataset~\cite{sun2019waymo}. Waymo does not include rear cameras' images, so we drop this input for UniAD-Single. VLM methods only use the front view and do not require any adaptation.

\smallskip
\boldparagraph{Results} As shown in \tabref{tab:arch}, UniAD-Single does not cope well with the new sensor configuration, and drops below BLIP-RT-2 in performance. The multi-stage approach of \algname provides further improvements. In particular, the accuracy of speed predictions rises from $43.90$ with no context to $54.29$ with the full graph. On the other hand, the chain approach does not provide sufficient useful information, with a speed accuracy of only $41.28$.

We present qualitative results for \algname on nuScenes and Waymo in~\figref{fig:qualitative}. The model generally provides intuitive answers, with a few exceptions (\textit{\eg}, planning on \dataname-nuScenes, perception on Waymo). This shows the utility of GVQA towards interactive driving systems. Further, on Waymo, we see meaningful prediction and planning answers despite the imperfect perception. For more visualizations, please see the supplementary material.

\subsection{{Question-wise Analysis in \dataname-nuScenes}}
\label{sec:exp_question_wise}
Next, we analyze the effect of each type of QA pair on the behavior performance by adding them as context. First, a set of representative QA pairs are selected in each stage. We then train BLIP-2 on different combinations of sets of QA pairs, and those QA pairs are added as the context for the behavior question (\tabref{tab:nus-question-wise}). 

\smallskip
\boldparagraph{Representative Questions at each stage}
We select different sets of representative QA pairs in each stage as follows:
\begin{itemize}[label=\raisebox{0.25ex}{\tiny$\bullet$},leftmargin=*]
\small
    \item $P_{1-1}$: What are the important objects in the current scene?
    \item $P_{1-2}$: What is the moving status of object \textit{X}?
    \item $P_{1-3}$: What is the visual description of object \textit{X}?
    \item $P_{2-1}$: What is the future state of object \textit{X}?
    \item $P_{2-2}$: Would object \textit{X} be in the moving direction of the ego vehicle?
    \item $P_{2-3}$: What object should the ego vehicle notice first / second / third when the ego vehicle is getting to the next possible location?
    \item $P_{3-1}$: What actions could the ego vehicle take based on the observation of object \textit{X}?
    \item $P_{3-2}$: What actions taken by the ego vehicle can lead to a collision with object \textit{X}?
    \item $P_{3-3}$: In this scenario, what are safe actions to take for the ego vehicle?
\end{itemize}

The reasons for such a selection are \textbf{1)} statistically they make up about 60\% of the total QA pairs in the three stages. \textbf{2)} subjectively they represent the most information needed for human to drive. A coarser stage-wise analysis is provided in the supplementary material for further investigation.

\smallskip
\boldparagraph{Results} Our results are shown in \tabref{tab:nus-question-wise}. We observe that training with QA pairs from prediction and planning stage (\textbf{ID} 4-9) improves the performance from training with perception QA pairs only (\textbf{ID} 1-3). Adding QA pairs from the planning stage (\textbf{ID} 7-9) does not significantly boost the performance compared to their previous stages (\textbf{ID} 4-6). The reason could be that other vehicles' future status contains all the necessary information to make the behavior decision.

\begin{table}[t!]
\caption{\textbf{{Question-wise analysis in \dataname-nuScenes.}} Questions of $P_{x-y}$ listed in \secref{sec:exp_question_wise}. Using Prediction and Planning stages of QA pairs as context for Behavior question improves the performance from using Perception only. However, the performance of different questions in the identical stage differentiate slightly.}
\vspace{-8pt}
\centering
\setlength{\tabcolsep}{5pt}
\resizebox{\textwidth}{!}{
\begin{tabular}{l|ccc|ccc|ccc|ccc}
\toprule
    \multirow{2}{*}{\textbf{ID}}& \multicolumn{3}{c|}{\textbf{Perception}} & \multicolumn{3}{c|}{\textbf{Prediction}} & \multicolumn{3}{c|}{\textbf{Planning}} & \multicolumn{3}{c}{\textbf{Behavior}}\\
    & $P_{1-1}$ & $P_{1-2}$ & $P_{1-3}$ & $P_{2-1}$ & $P_{2-2}$ & $P_{2-3}$ & $P_{3-1}$ & $P_{3-2}$ & $P_{3-3}$ & Acc. $\uparrow$ & Speed $\uparrow$  & Steer $\uparrow$  \\
    \midrule
    1 & \checkmark & - & - & - & - & - & - & - & - & 54.69 & 66.83 & 75.22 \\
    2 & \checkmark & \checkmark & - & - & - & - & - & - & - & 55.32 & 67.33 & 74.34 \\
    3 & \checkmark & \checkmark & \checkmark & - & - & - & - & - & - & 53.94 & 65.33 & 75.72 \\
    \midrule
    4 & \checkmark & \checkmark & \checkmark & \checkmark & - & - & - & - & - & {58.82} & 71.83 & \textbf{80.98} \\
    5 & \checkmark & \checkmark & \checkmark & \checkmark & \checkmark & - & - & - & - & 57.07 & 71.96 & 78.97 \\
    6 & \checkmark & \checkmark & \checkmark & \checkmark & \checkmark & \checkmark & - & - & - & 57.70 & 72.22 & 79.22 \\
    \midrule
    7 & \checkmark & \checkmark & \checkmark & \checkmark & \checkmark & \checkmark & \checkmark & - & - & \textbf{58.95} & 72.59 & 80.23 \\
    8 & \checkmark & \checkmark & \checkmark & \checkmark & \checkmark & \checkmark & \checkmark & \checkmark & - & 57.95 & \textbf{72.97} & 79.97 \\
    9 & \checkmark & \checkmark & \checkmark & \checkmark & \checkmark & \checkmark & \checkmark & \checkmark & \checkmark & 57.49 & 69.89 & 80.63 \\
    
\bottomrule
\end{tabular}}
\label{tab:nus-question-wise}
\vspace{-0.4cm}
\end{table}

\subsection{Performance for $P_{1-3}$ via GVQA}
\label{sec:exp_vqa}

In our final experiment, we establish baseline results for the $P_{1-3}$ stages of GVQA, studying the impact of context. We use two VLMs, the off-the-shelf BLIP-2~\cite{li2023blip2} (not fine-tuned on \dataname), and the proposed \algname. 

\smallskip
\boldparagraph{Baselines} We consider the lower bound of no context (`None'), which corresponds to training and evaluation with the same setting as standard VQA (image and question in, answer out). As an upper bound for each architecture, we perform GVQA but input the ground truth (`GT') context to the model at test time instead of its own prior predictions.

\smallskip
\boldparagraph{Results} Our results are summarized in \tabref{tab:vqa}. It can be observed that \dataname-nuScenes is more challenging for both models, as indicated by the lower scores on it relative to \dataname-CARLA in all context settings. This is likely due to the higher diversity in human answers obtained for \dataname-nuScenes, as opposed to the rule-based generation in CARLA. On both datasets, \algname, which is fine-tuned on \dataname, significantly outperforms BLIP-2 which is applied in a zero-shot manner. 
{
On both datasets and cases (zero-shot and fine-tuned), we see the potential of the graph-based structure.
{For the exact evaluation setting and a question-wise performance analysis, we refer to the supplementary.}
}

\begin{table}[t!]
\caption{\textbf{Baseline $P_{1-3}$ Results.}
{\algname and the zero-shot BLIP-2 benefit from a step-wise reasoning procedure given by our graph structure.
}}
\vspace{-8pt}
\centering
\setlength{\tabcolsep}{5pt}
\resizebox{\textwidth}{!}
{
\begin{tabular}{l|cc|cc|cc|cc}
\toprule
\multirow{3}{*}{\textbf{Context}} & \multicolumn{4}{c|}{\textbf{\dataname-nuScenes} ($P_{1-3}$)}  & \multicolumn{4}{c}{\textbf{\dataname-CARLA} ($P_{1-3}$)}  \\
\cmidrule{2-9}
& \multicolumn{2}{c|}{BLIP-2~\cite{li2023blip2}} & \multicolumn{2}{c|}{\algname} & \multicolumn{2}{c|}{BLIP-2~\cite{li2023blip2}} & \multicolumn{2}{c}{\algname} \\
& SPICE $\uparrow$ & GPT $\uparrow$ & SPICE $\uparrow$ & GPT $\uparrow$ & SPICE $\uparrow$ & GPT $\uparrow$ & SPICE $\uparrow$ & GPT $\uparrow$ \\
\midrule
None & 4.34 & 42.97 & 42.56 & 71.39 & {10.46}  & {46.37} & {72.71} &  {79.67} \\
Graph & 7.71 & 45.21 & 49.54 & 72.51 & {10.30} & {55.03} & {75.26} & {81.78}\\
\midrule
\textit{GT} & \textit{8.19}  & \textit{41.10}& \textit{50.29} & \textit{72.94} & \textit{{16.18}} & \textit{{57.98}} & \textit{{79.07}} & \textit{{83.13}} \\
\bottomrule
\end{tabular}}
\label{tab:vqa}
\vspace{-0.4cm}
\end{table}

\section{Related Work}
\vspace{-4pt}

\boldparagraph{Generalization in Autonomous Driving} The inadequacy of generalization to the ``long tail'' of corner cases poses significant safety concerns to AD systems~\cite{chen2023end,teng2023motion,tampuu2020survey}. To tackle this issue, prior research primarily makes efforts in data-driven methods~\cite{suo2021trafficsim,chen2022learning,akhauri2020enhanced,wang2021advsim,hanselmann2022king}. For example, TrafficSim~\cite{suo2021trafficsim} collects more data for safety-critical cases by simulation. An emerging direction involves leveraging semantic information to supervise the detection of unseen or anomalous objects~\cite{elhafsi2023semantic,wang2023driveanywhere,chen2023categorical,pan2024vlp}. 
Even so, the zero-shot performance of AD systems is currently not satisfactory. In this paper, we bring a new approach towards better generalization: learning logical reasoning using Graph VQA.

\smallskip
\boldparagraph{Language-grounded Driving} Several concurrent methods attempt to incorporate multi-modal inputs into LLMs for AD tasks~\cite{seff2023motionlm,mao2023gptdriver,chen2023drivingwithllms,jin2023adapt,xu2023drivegpt4,wayve2023lingo1,wang2023driveanywhere,marcu2023lingoqa,chen2023categorical,wang2023drivemlm,mao2023language,tian2024drivevlm,shao2023lmdrive,yuan2024ragdriver,pan2024vlp,han2024dmedriver}. Specifically, GPT-Driver~\cite{mao2023gptdriver} and LLM-Driver~\cite{chen2023drivingwithllms} encode the perceived scene state into prompts, relying on LLMs to formulate reasonable plans. DriveGPT4~\cite{xu2023drivegpt4} projects raw sensor data into tokens and utilizes LLMs for end-to-end prediction of control signals and explanations. Despite these preliminary attempts, there is untapped potential in addressing generalization in AD through LLMs. Our work combines VLMs with training over graph-structured QAs from \dataname. This enables us to show benefits on zero-shot end-to-end planning, which was not demonstrated by these concurrent studies.
\section{Discussion}
\label{sec:discussion}
\vspace{-4pt}

\begin{wraptable}{r}{6cm}
\vspace{-1.2cm}
\caption{{Compared to UniAD-Single, \algname has fewer trainable parameters but a higher inference cost.}}
\centering
\small
\resizebox{0.5\textwidth}{!}{
\begin{tabular}{l|cccc}
\toprule   
    Method  &\#Params& \#Trainable & FLOPs &   FPS    \\ \midrule
    UniAD-Single & 131.9M & 58.8M & 1.7T & 1.8  \\
    \rowcolor[gray]{0.9} \algname & 3.955B & 12.9M & 24.2T& 0.16\\
\bottomrule
\end{tabular}
}
\vspace{-0.5cm}
\label{tab:latency}
\end{wraptable}

Even though DriveLM exhibits promising generalization, there are concerning limitations of this work.

\begin{itemize}
    \item \textbf{Efficiency Constraints.} Inheriting the drawbacks of LLMs, our baseline model suffers from long inference times, especially as we require multiple rounds of predictions based on the graph structure (roughly $10\times$ slower than UniAD, {as shown in \tabref{tab:latency}). The core problem lies in the slow throughput of the current model (\textbf{8.5 tokens/s} in \algname),} which may impact practical usage. {However, LLMs have become orders of magnitude faster~\cite{xiao2023smoothquant,shi2023crossget} in the last months as this is a general topic of broad interest. We believe that the rapid progress in orthogonal research can alleviate this issue.}
    \vspace{-0.3cm}
\end{itemize}

\begin{itemize}
    \item \textbf{Driving-specific Inputs.} \algname directly applies the VLM's vision module, taking a low-resolution front-view image as input. Driving-specific sensors such as LiDAR cannot be processed as well. This results in a lack of temporal information and 360-degree scene understanding. Extending \algname to images from multiple views is straightforward as the graph formulation allows various input frames for different nodes. We leave it for future work to explore options for multi-modal and multi-frame inputs.
    \item \textbf{Closed-loop Planning.} Our approach is currently evaluated under an open-loop scheme. In this setting, incorporating the ego vehicle's status as input can significantly enhance the metrics, but its effectiveness may not translate well to the real world, and hence we only consider methods that do not do so. Extending our work to a closed-loop setting with an affordable budget in training time and computational cost is a promising direction to explore. With the usage of CARLA we provide a promising foundation for more research in the direction of closed-loop planning with VLMs.
    \vspace{-0.3cm}
\end{itemize}

\smallskip
\boldparagraph{Conclusion} We show how VLMs can be leveraged as end-to-end autonomous driving agents with improved generalization over task-specific driving stacks. For this we propose the task of Graph VQA together with new datasets and metrics. Equipped with these tools, we build a baseline approach that has a simple architecture and obtains promising results.

\section*{Acknowledgements}
The OpenDriveLab team is part of the Shanghai AI Lab and kindly supported by National Key R\&D Program of China (2022ZD0160104) and NSFC (62206172).
This paper is partially supported by the National Key R\&D Program of China No.2022ZD0161000 and the General Research Fund of Hong Kong No.17200622 and 17209324. 
This work was also supported by the BMBF (Tübingen AI Center, FKZ: 01IS18039A), the DFG (SFB 1233, TP 17, project number: 276693517), and by EXC (number 2064/1 – project number 390727645). We thank the International Max Planck Research School for Intelligent Systems (IMPRS-IS) for supporting K. Renz and K. Chitta.
Our gratitude goes to Tai Wang for the valuable feedback,
Qingwen Bu for refining the figures,
and Jiazhi Yang, Shenyuan Gao, Yihang Qiu, Tianyu Li, Yunsong Zhou, Zetong Yang, Julian Zimmerlin for the fruitful discussion.

\bibliographystyle{splncs04}
\bibliography{bibliography_short, bibliography_custom, bibliography}

\newpage

\section*{Overview}
\medskip
\noindent In the appendices below, we first delve deeper into various discussions, along with additional details around the annotation process of \dataname-nuScenes \& -CARLA, GVQA metrics, context setting \& trajectory tokenization in \algname, and more ablation results of \algname on \dataname-nuScenes and Waymo.
Finally, we provide additional results and visualizations that further complement the findings from the main text.

For readers who want to focus on specific topics, we provide a summary below:

\medskip
\noindent \rule{\linewidth}{0.5pt}
\medskip

\appendix

\noindent \hyperref[sec:motivating_questions]{Appendix A -- \textit{Motivating Questions}}

\begin{quote}
    \medskip
    We index a list of ``motivating'' questions that may arise from reading the main text and that we expand on further here (\textit{e.g.}, ``why adapt general VLMs to driving''). These questions are open to be explored and thus our answers here are \textit{intuitive} and \textit{empirical}.
\end{quote}

\bigskip

\noindent \hyperref[supp:nuScenes]{Appendix B -- \textit{\dataname-nuScenes}}

\begin{quote}
    \medskip
    We provide the DriveLM-nuScenes dataset composition, introduce the detailed annotation pipeline and conduct statistics of QA categories.

\end{quote}

\bigskip

\noindent \hyperref[supp:CARLA]{Appendix C -- \textit{\dataname-CARLA}}

\begin{quote}
    \medskip
    We provide a detailed description of the PDM-Lite expert, composition of the dataset, how the graph looks, and explain the data generation and annotation process.

\end{quote}

\bigskip

\noindent \hyperref[supp:metrics]{Appendix D -- \textit{\dataname-Metrics}}

\begin{quote}
    \medskip
    We explain the details of the metrics for each task in the GVQA, illustrate their differences, and provide the reasons for proposing GPT-score as the main metric used in the $P_{1-3}$ VQA tasks.

\end{quote}

\bigskip

\noindent \hyperref[supp:agent]{Appendix E -- \textit{\algname}}

\begin{quote}
    \medskip
    We introduce the detailed design of the prompting with context and the trajectory tokenization, including the differences in context during training and evaluation, the pattern for the trajectory as a sentence, and the hyperparameters in the tokenizer.
\end{quote}

\bigskip

\noindent \hyperref[supp:exp]{Appendix F -- \textit{Experiments}}

\begin{quote}
    \medskip
    We provide more experiments, including the generalization to unseen objects, effects of more context design on the zero-shot ability, the performance with more conventional VQA metrics and the model efficiency comparison.
\end{quote}

\bigskip

\noindent \hyperref[supp:qualitative]{Appendix G -- \textit{Qualitative Results}}

\begin{quote}
    \medskip
    We show qualitative examples of the context, questions, and answers on nuScenes, Waymo, and CARLA. Additionally, we contrast predicted and ground truth answers together with their SPICE and GPT Score on nuScenes to provide some intuition for those metrics.
\end{quote}

\bigskip

\noindent \hyperref[supp:related]{Appendix H -- \textit{Other Related Work}}

\begin{quote}
    \medskip
    We provide more related works from two new perspectives. One is reasoning over graph structure which is similar to our idea of graph-structure reasoning, the other is more vision-language benchmarks for autonomous driving.
\end{quote}

\bigskip

\clearpage
\newpage

\clearpage
\newpage

\section{Motivating Questions}
\label{sec:motivating_questions}

\bigskip
\noindent\textbf{Q1.} \textit{In what situations could we expect planning with VLMs to outperform conventional end-to-end autonomous driving?}

\smallskip
One of the key challenges of autonomous driving is to generalize to the long-tail of scenarios, that are rarely encountered but have critical importance.
Considering the large-scale pre-training of VLMs, their acquired knowledge of the world, and the reasoning ability of the LLM, it is anticipated that planning with VLMs will work better in situations that are novel or unseen in the context of driving scenarios but encountered during pre-training in unrelated contexts.

\bigskip
\noindent\textbf{Q2.} \textit{Why adapt general VLMs to driving rather than adding language inputs to driving-specific models?}

\smallskip
General VLMs benefit from billion-scale pre-training data for vision-language tasks extracted from the internet, which can be adapted to the driving domain through fine-tuning on small autonomous driving datasets like \dataname. Conversely, driving-specific models are only pre-trained on small autonomous driving datasets, and adding language inputs to these with data from outside the self-driving domain is non-trivial. Combining the advantages of VLMs and driving-specific models is however an interesting direction to explore.

\bigskip
\noindent\textbf{Q3.} \textit{Can open-loop evaluation of planning provide meaningful results?}

\smallskip
When performing open-loop evaluation, providing the ego history as an input to the planning module prevents fair comparisons, as this signal alone is sufficient for achieving low errors on existing benchmarks. \dataname alleviates this issue by evaluating key frames, where the intention of the ego vehicle changes, and the ego history is not strongly indicative of the future behavior or motion. Additionally, we consider baselines in our analysis that do not input the ego history to the planning module. Finally, we introduce \dataname-CARLA as a means to show closed-loop planning results in the future.

\bigskip
\noindent\textbf{Q4.} \textit{Why are there currently no closed-loop planning results on CARLA?}

\smallskip
Running 4B parameter models at 20 FPS as required by CARLA needs more engineering effort. This could be solved by using distillation, quantization, and caching techniques in LLM inference. Another approach would be to execute only the final motion stage of \algname at 20 FPS, while the other GVQA stages are executed at a lower frame rate.

\bigskip
\noindent\textbf{Q5.} \textit{Is \algname efficient enough to be applicable to real-world autonomous driving?}

\smallskip
We comment on the runtime of \algname in Table~\textcolor{red}{5} in the main paper. Without any optimization, the approach is around 1 order of magnitude slower than UniAD. However, with the optimizations proposed for closed-loop results on CARLA (see \textbf{Q4}), practical applications of VLMs in driving should be possible. 

\bigskip
\noindent\textbf{Q6. } \textit{What is the trade-off between long inference time and generalization?}

\smallskip
LLMs have become orders of magnitude faster in the last years as they are a general topic of broad interest, {e.g. \cite{xiao2023smoothquant}}. Similarly, recent works show BLIP-2 can be run 40\% faster while maintaining performance \cite{shi2023crossget}.
Furthermore, most AD research begins with systems that are not real-time, which are later optimized by practitioners. We admit that runtime efficiency is a limitation of DriveLM but we hope that the rapid progress in orthogonal research can alleviate this issue, which is out of the scope of our project.

\bigskip
\noindent\textbf{Q7.} \textit{Why is VQA more suited than alternative techniques to train internet-scale models (such as generative modeling) for the downstream application of autonomous driving?}

\smallskip
Both perception and planning in driving require reasoning and involve zero-shot generalization. VLMs potentially have the reasoning ability inherited from LLMs, making VQA a promising direction for bringing the benefits of web-scale training to autonomous driving.

\bigskip
\noindent\textbf{Q8.} \textit{Do today's VLMs understand and reason about the visual world as well as LLMs understand text-based worlds?}

\smallskip
This is not known but deserves to be explored. VLMs approach the problem of generalization in a data-driven way, which has been proved successful repeatedly on other tasks.

\bigskip
\noindent\textbf{Q9.} \textit{Why does the proposed graph reasoning scheme not provide very strong improvements in VQA?}

\smallskip
It is possible that the simple prompting scheme, relatively small base VLMs, or insufficiently strong logical dependencies in the dataset (or a combination of these factors) contribute to the lack of major improvements. \dataname-CARLA provides a platform to carefully study these factors and inform the annotation of future datasets for GVQA.

\bigskip
\noindent\textbf{Q10.} \textit{What questions should be asked when collecting DriveLM?}

\smallskip
Determining the right questions is a critical aspect of the DriveLM system. 
We are unable to recommend a detailed protocol for this in our work, which is a pioneering study for driving with VLMs. 
This relies on domain expertise, which we aim to address in future work.
See Table \textcolor{red}{3} in the main paper for a question-wise analysis of our chosen protocol.

\bigskip
\noindent\textbf{Q11. } \textit{Is generalization in automobiles an oversold topic?}

\smallskip
Driving scenarios predominantly consist of vehicles, pedestrians, and cyclists. 
However, real-world AD is a safety-critical application that must handle the long tail of rare objects to avoid accidents in commercial deployment \cite{2022incidentai}.
Furthermore, generalization to novel objects is an active and growing research field as shown in \cite{zareian2021openvocabulary,li2022coda}.
While it is true that most of the objects in autonomous driving are limited, the tiny percentage of rare objects are in fact the main existing barrier to the commercialization of AD, making generalization an important feature.

\bigskip
\noindent\textbf{Q12. } \textit{Why not use video inputs?}

\smallskip
It has been observed that in AD multi-frame inputs do not always lead to improved performance due to causal confusion \cite{wen2021keyframefocused}, motivating the use of single frames in several prior works \cite{wu2022trajectoryguided,jaeger2023hidden}.
However, we agree that in the long term, multi-frame models are desirable. As a proof of concept, we provide the results for one (adapted from LLaMA-Adapter-V2 \cite{gao2023llamaadapterv2}) in \tabref{tab:llama}. Multi-frame inputs give a slight improvement.

\bigskip
\noindent\textbf{Q13. } \textit{What is the technical novelty in this paper?}

\smallskip
The young field of driving with language still lacks a standardized dataset, task and evaluation framework. Our paper fills this gap. We believe that these are as valuable as proposing a new method. Therefore, we do not make any claims regarding algorithmic novelty, intentionally constructing simple baselines. Our technical novelty lies in the formulation of the task and preparation of data suitable for adapting general VLMs to the driving application, which we show to have promising results even in combination with existing models.

\section{DriveLM-nuScenes}
\label{supp:nuScenes}

In this section, we introduce the details of \dataname-nuScenes, including the dataset composition, collection methodology, and statistics.

\subsection{Dataset Composition}
\label{supp:nuScenes-data-comp}

\dataname-nuScenes comprises \textbf{a training set of 4072 frames and a validation set of 799 frames}, consisting of scene-level descriptions and frame-level QA accompanied by 2D bounding boxes within multi-view images from the nuScenes dataset.
The scene-level description delineates the behavior of the ego vehicle throughout the entire video clip.
The frame-level QA encompasses three distinct categories: perception, prediction, and planning.

\begin{itemize}[labelindent=16pt]
    \item \textbf{Perception} involves queries related to the thorough examination of the entire frame. Apart from several questions in this question set that are manually annotated, we design prompts to generate questions about the observational facets of objects within the scene, leveraging ground truth from nuScenes~\cite{caesar2019nuscenes} and OpenLane-V2~\cite{wang2023openlanev2}.

    \item \textbf{Prediction} encompasses a series of inquiries regarding the projection of the forthcoming state of key objects and the ego vehicle in the current frame, and the underlying reasoning process behind the prediction.
    Because the predictions are intricate and challenging, we \textit{manually annotate} the answers.

    \item \textbf{Planning} contains questions related to planning subsequent actions of the ego vehicle in the current frame. As ``Planning'' is the same challenging as ``prediction'', we design the prompt for the reasoning process and \textit{manually annotate} the answers to the questions.

\end{itemize}

\begin{figure*}
  \centering
  \includegraphics[width=\linewidth]{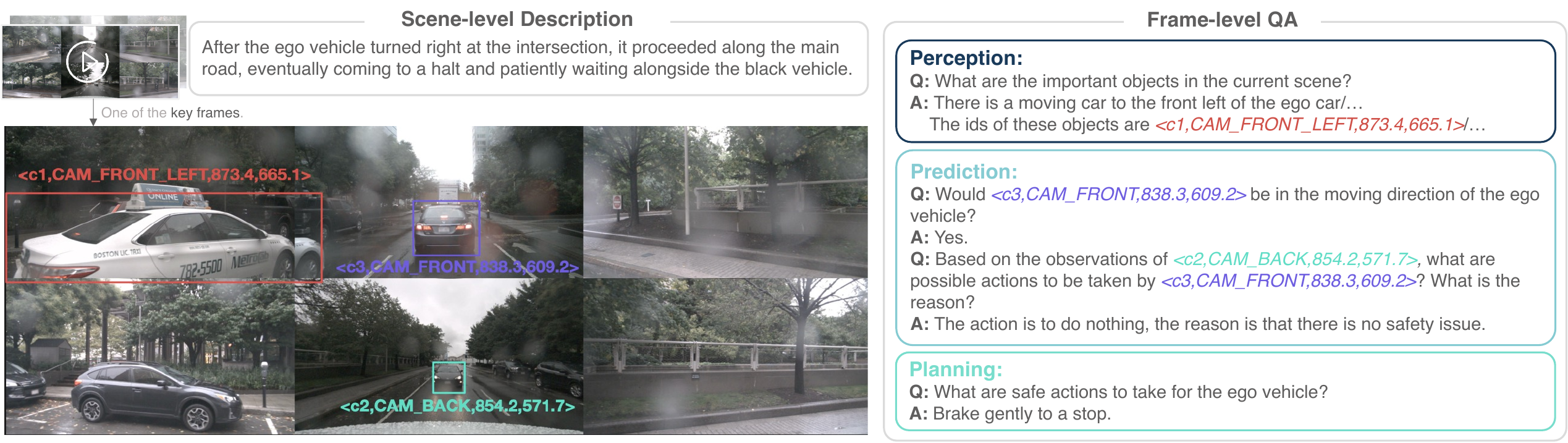}
  \caption{
  \textbf{Overall Composition of \dataname-nuScenes}. 
  The dataset comprises scene-level descriptions and frame-level QA, which can be divided into three parts: \textit{Perception}, \textit{Prediction}, and \textit{Planning}. 
  Objects are encoded using \textit{c tags}, which contain identifiers, camera affiliations, and center coordinates of its 2D bounding box in the corresponding camera frame.
  }
  \label{fig:composition}
\end{figure*}

For the key objects referred to in the QA, we encode them as \textbf{c tags} in the format <$c$, $CAM$, $x$, $y$>, where $c$ is the identifier, $CAM$ indicates the camera where the object's center point is situated, and $x$, $y$ represent the horizontal and vertical coordinates of the 2D bounding box in the respective camera's coordinate system.
We also provide a dictionary in each key frame, recording more basic information about the key objects such as the size of the bounding box, the category, the moving state, and the visual description.
The overview of data organization forms is shown in Fig.~\ref{fig:composition}.

\subsection{Collection Methodology}
\label{supp:nuScenes-coll-method}

\begin{figure*}
  \centering
  \includegraphics[width=\linewidth]{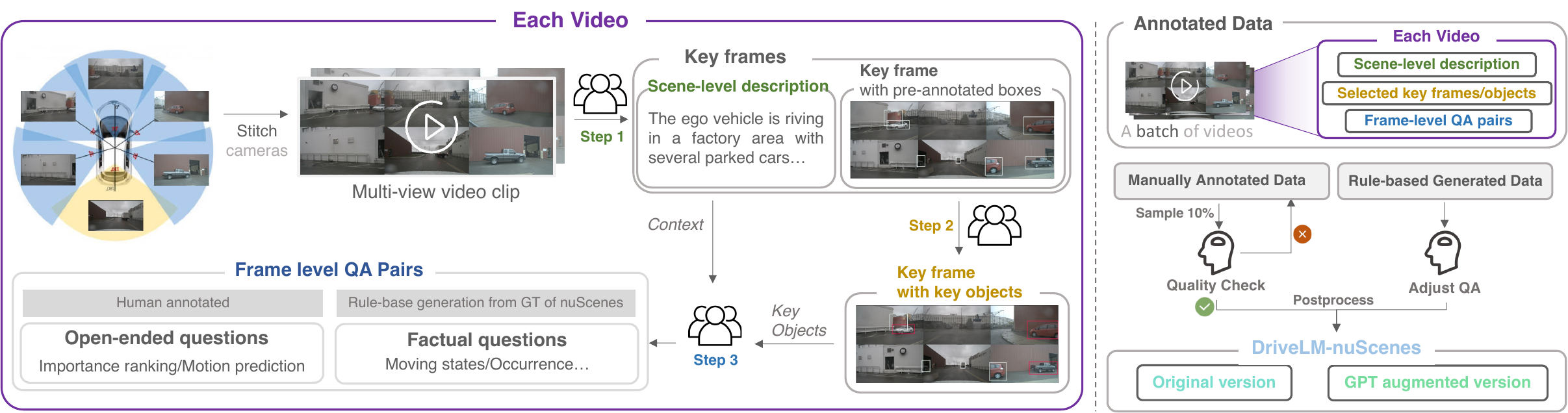}
  \caption{\textbf{(Left) Pipeline of the three-steps annotation process.} For each video, we ask the annotators to annotate the key frames, key objects, and QA attributes step by step. \textbf{(Right) The quality check and post-processing progress.} We divided the annotated data into batches, where each batch contains 8 video clips and their related annotations. We conduct rigorous quality checks, and after the post-processing, we finally get two versions of our \dataname-nuScenes dataset.}
  \label{fig:nus_anno_pipeline}
\end{figure*}

During the annotation process, we employ individuals with driving experience for the labeling task. 
We provide annotators with the stitched results from the six cameras of nuScenes as source data.
As shown in Fig.~\ref{fig:nus_anno_pipeline} (left), we divide the annotation process into three steps: selecting key frames from video clips, choosing key objects within these key frames, and subsequently annotating the frame-level QAs in the key frames. 
Following this, we conduct multiple rounds of quality checks to ensure the data reliability and perform post-processing procedures on the qualified data as shown in Fig.~\ref{fig:nus_anno_pipeline} (right).
The specific details of this pipeline will be introduced below.

\bheading{Key Frame Selection.}
In this process, we ask annotators to review the entire video clip to pinpoint key frames rich in scene information and potentially indicative of future state changes. 
Simultaneously, annotators are instructed to label the ego vehicle's behavior throughout the video clip. This segment serves as the foundation for our scene-level description.

\bheading{Key Object Selection.}
In this annotation step, we instruct annotators to identify objects in key frames that are relevant to the ego vehicle's driving, denoted as key objects. 
To ensure accuracy, we provide pre-annotated bounding boxes based on ground truth categories from nuScenes~\cite{caesar2019nuscenes}. 
Annotators also have the flexibility to designate objects not present in the ground truth as key objects if they are deemed significant. 

\bheading{QA Labeling.}
In the QA labeling process, we have two sets of questions, factual questions and open-ended questions.
For the factual questions, we generate the answers with a rule-based method.
For the open-ended questions, we instruct annotators to manually annotate the meticulously designed questions. 
Options are provided for most manually annotated questions, and
we include an "Other - Fill in the Blank" option for answer choices in such cases to ensure flexibility. 
We have also incorporated free-form questions, allowing annotators to generate their own inquiries about the current frame.

\bheading{Quality Check.}
We prioritize the quality of our data. 
In addition to establishing clear criteria and implementing autonomic checking strategies at each annotation step, we conduct rigorous manual quality checks.
We organize the final data into batches, with each batch comprising 8 video clips, along with their scene-level descriptions, key frames with key objects selected from the 8 video clips, and corresponding QA pairs for each key frame. 
We provide explicit standards to quality check inspectors, instructing them to assess data eligibility based on these criteria. 
For manually annotated data, if the accuracy of the manual annotations falls below expectations for a particular batch, we compile feedback on the issues encountered and request annotators to re-annotate the entire batch.
For data generated from ground truth, we instruct quality inspectors to manually adjust the inconsistent or unreasonable QA pairs.

\bheading{Post Processing.}
Since our annotators are Chinese speakers, we need to translate the labeled data into English after obtaining it. 
Initially, we establish mappings between Chinese and English using a vocabulary. 
For texts that are not successfully mapped, we utilize GPT-3.5 
for translation, and then perform manual checks and corrections on the GPT outputs.
We also provide a version augmented by GPT-3.5, utilizing the prompt as shown in \tabref{box:refine}.
{
\captionsetup{type=table}
\begin{tcolorbox}[colback=gray!10,
                  colframe=black,
                  width=\linewidth,
                  arc=1mm, auto outer arc,
                  boxrule=0.5pt,
                 ]

{\cmtt{\textcolor{blue}{\textbf{Messages}}
 = [ 

\smallskip
 \{
"role": "system",
"content": f"""} You are an English improver.
\cmtt{""" \},}

\smallskip
\cmtt{\{
"role": "user",
"content": f"""} I have a question and answer that I need you to help me modify and embellish, please make a few simple changes to the content in written language and keep the meaning same, you only need to answer the changes to: 
\{\cmtt{\textcolor{blue}{QA}\}"""}\}]}

\end{tcolorbox}
\vspace{-10pt} \captionof{table}{\textbf{Prompt for GPT-refined version of \dataname-nuScenes}. We try 50 different prompt and select this pattern as the final one to do the refinement.}\label{box:refine}
}

\begin{figure*}
    \centering
    \includegraphics[width=0.45\linewidth]{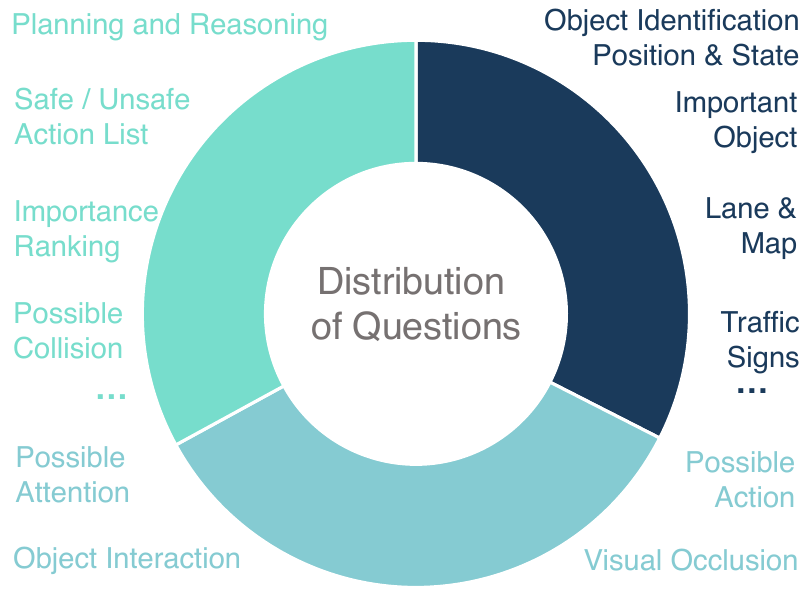}
    \caption{ \textbf{Question Distribution of} \textbf{\textcolor[rgb]{0.12549019607843137,0.19607843137254902,0.3137254901960784}{Perception}}, \textbf{\textcolor[rgb]{0.5254901960784314,0.7607843137254902,0.796078431372549}{Prediction}} and \textbf{\textcolor[rgb]{0.49411764705882355,0.8392156862745098,0.7686274509803922}{Planning}:} The questions in our dataset cover various specific aspects of driving tasks, generally categorized into perception, prediction, and planning. Most of these questions are annotated by human annotators, making this a suitable proxy for human-like driving reasoning.}
    \label{fig:data_distribution}
\end{figure*}

\subsection{Statistics and Facts}
\label{supp:nuScenes-stats}
In this section, we conduct a distribution analysis of our DriveLM-nuScenes QA categories at both the task level and object level. 
Additionally, for the task level, we provide the templates for all our QA under this classification criterion.
The results indicate the richness of our QA categories, covering various aspects of autonomous driving.
Moreover, the abundance of logical relationships is sufficient to construct a graph-structured QA.
\begin{figure*}
  \centering
  \includegraphics[width=\linewidth]{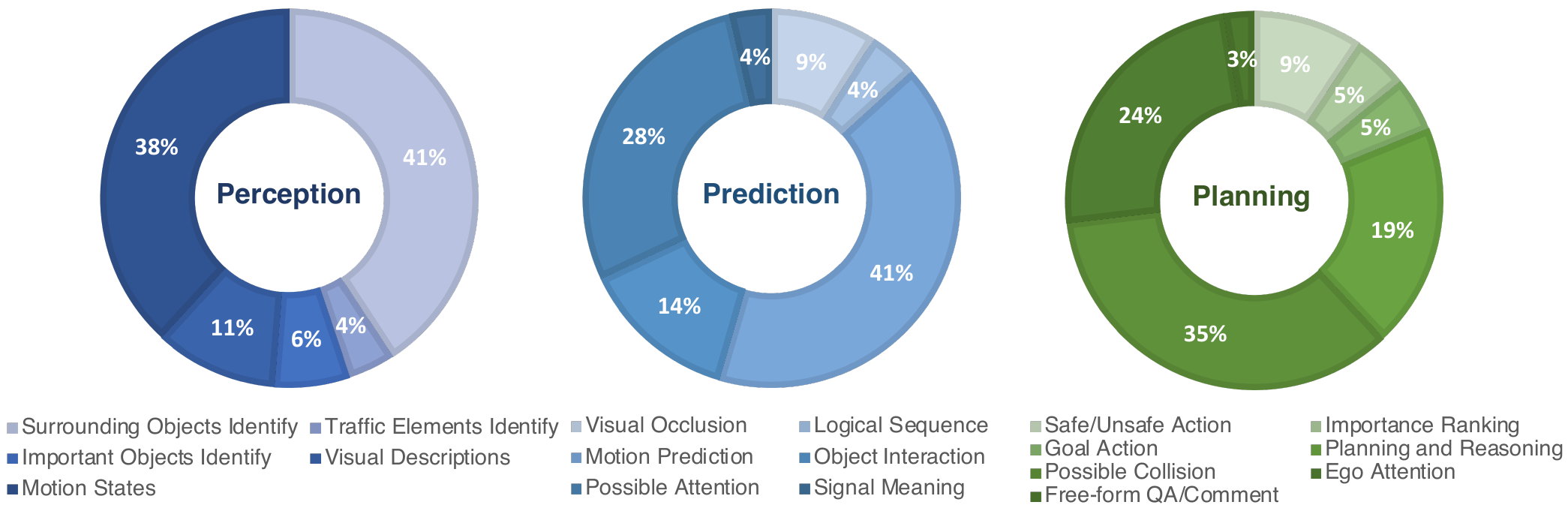}
  \caption{\textbf{The distribution of question types according to different tasks in \dataname-nuScenes.} We categorize questions into perception, prediction, and planning tasks, each further subdivided into more specific question types.}
  \label{fig:nus_statis}
\end{figure*}

\begin{figure}[]
  \centering
  \includegraphics[width=\linewidth]{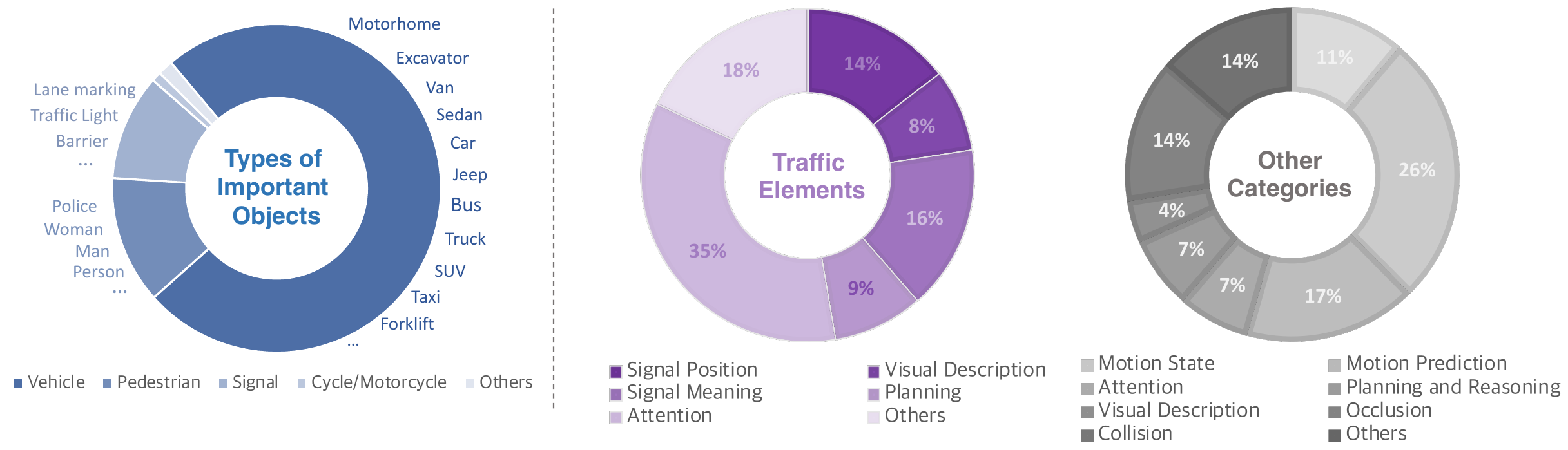}
  \caption{\textbf{(Left) The distribution of key objects in \dataname-nuScenes.} The sub-categories are extracted from the visual description. \textbf{(Right) The distribution of question types related to different key objects in \dataname-nuScenes.} Since the questions associated with traffic elements differ significantly from other categories, we separately conduct statistics for QA types related to traffic elements and the remaining categories. }
  \label{fig:nus_obj_stats}
\end{figure}

\bheading{Task level.} Our \dataname-nuScenes orchestrates a benchmark that encompasses various aspects of autonomous driving, connecting the whole stages of the human driving logic.
To delve deeper into this aspect, we present the detailed QA types distribution at task level in~\figref{fig:nus_statis}. 
For a better understanding, we also provide examples of QA templates in all of the P3 stages in \tabref{box:nus_qa}.
{
\captionsetup{type=table}
\begin{tcolorbox}[colback=gray!10,
                  colframe=black,
                  width=\linewidth,
                  arc=1mm, auto outer arc,
                  boxrule=0.5pt,
                  breakable
                 ]

\textcolor{blue}{\textbf{Perception}}
\smallskip

\textcolor{blue}{Surrounding Objects Identify}
\smallskip

Q: Please describe the current scene.

A: There are two moving cars behind the ego car and two barriers in front of it.
\smallskip

Q: What are objects to the front left/back right/... of the ego car?

A: There are two barriers to the front left of the ego car.
\smallskip

Q: Are there traffic cones/moving cars/... to the front right/back left/... of the ego car?

A: No.
\smallskip

\textcolor{blue}{Traffic Elements Identify}
\smallskip

Q: Is there any traffic element in the front view?

A: Yes, there are some traffic elements in the front view.
\smallskip

Q: Identify all the traffic elements in the front view, categorize them, determine their status, and predict the bounding box around each one. The output should be a list formatted as (c, s, x1, y1, x2, y2), where c represents the category, s denotes the status, and x1, y1, x2, y2 are the offsets of the top-left and bottom-right corners of the box relative to the center point.

A: There are three traffic elements in the front view. The information of these traffic elements are [(road sign, go straight, 907.58, 590.67, 992.54, 630.95)...].
\smallskip

\textcolor{blue}{Important Objects Identify}
\smallskip

Q: What are the important objects in the current scene? Those objects will be considered for the future reasoning and driving decision.

A: There is a parked truck to the back of the ego car... The ids of these objects are <c1,CAM\_BACK,827.5,484.2>...
\smallskip

Q: What is the relative positioning of the important objects in the current scene?

A: <c3,CAM\_FRONT,689.2,527.5> is to the front of <c1,CAM\_BACK,827.5,484.2>...
\smallskip

Q: Which lanes are each important object on in the scene?

A: <c2,CAM\_FRONT,820.8,473.3> is on the ego lane...
\smallskip

\textcolor{blue}{Visual Description}
\smallskip

Q: What is the visual description of <c2,CAM\_FRONT\_LEFT,415.8,580.8>/...?

A: Pedestrian riding a bicycle.
\smallskip

\textcolor{blue}{Motion State}
\smallskip

Q: What is the status of the cars/pedestrians/... that are to the front/front right/... of the ego car?

A: Many cars are parked.
\smallskip

Q: What is the observed status of object <c1,CAM\_FRONT,920.0,509.2>/...?

A: Moving.
\smallskip

Q: What is the moving status of object <c1,CAM\_FRONT,920.0,509.2>/...?

A: Going ahead.
\smallskip

\textcolor{blue}{\textbf{Prediction}}
\smallskip

\textcolor{blue}{Visual Occlusion}
\smallskip

Q: Which object is most likely to be occluded by <c1,CAM\_FRONT,707.5,472.5>/...? Would this object affect the ego vehicle? Based on this object, what action of the ego vehicle is dangerous?

A: The object in front of <c1,CAM\_FRONT,840.8,507.5>, yes, accelerating forward.
\smallskip

\textcolor{blue}{Logical Sequence} 
\smallskip

Q: What object should the ego vehicle notice first when the ego vehicle is getting to the next possible location? What is the state of the object that is first noticed by the ego vehicle and what action should the ego vehicle take? What object should the ego vehicle notice second when the ego vehicle is getting to the next possible location? What is the state of the object perceived by the ego vehicle as second and what action should the ego vehicle take? What object should the ego vehicle notice third? What is the state of the object perceived by the ego vehicle as third and what action should the ego vehicle take?

A: Firstly notice that <c2,CAM\_FRONT,514.7,462.2>, the state of it is traffic sign, so the ego vehicle should slow down and go ahead. Secondly notice that <c3,CAM\_FRONT,950.3,613.1>, the state of it is traffic sign, so the ego vehicle should slow down and go ahead. Thirdly notice that <c1,CAM\_FRONT,707.5,472.5>, the state of it is going ahead, so the ego vehicle should slow down and go ahead.
\smallskip

\textcolor{blue}{Motion Prediction} 
\smallskip

Q: Would <c1,CAM\_FRONT,920.0,509.2>/... be in the moving direction of the ego vehicle?

A: Yes.
\smallskip

Q: What is the future state of <c1,CAM\_FRONT,920.0,509.2>/...?

A: Keep going straight.
\smallskip

Q: Will <c2,CAM\_FRONT,1223.3,598.3>/... be in the moving direction of <c1,CAM\_BACK,514.2,503.3>/...?

A: No.
\smallskip

\textcolor{blue}{Object Interaction}
\smallskip

Q: Will <c2,CAM\_FRONT,1223.3,598.3>/... change its motion state based on <c1,CAM\_BACK,514.2,503.3>/...?

A: No.
\smallskip

Q: Based on the observations of <c1,CAM\_BACK,514.2,503.3>/..., what are possible actions to be taken by <c2,CAM\_FRONT,1223.3,598.3>/...? What is the reason?

A: The action is to keep going at the same speed, the reason is there is no safety issue.
\smallskip

Q: Based on the observation of <c4,CAM\_FRONT,1071.2,346.2>/..., what actions may <c1,CAM\_FRONT,1126.7,515.0>/... take?

A: The action is to keep going at the same speed, the reason is there is no safety issue.
\smallskip

\textcolor{blue}{Possible Attention}
\smallskip

Q: In this scenario, what object is most likely to consider <c3,CAM\_FRONT,400.1,717.2>/...?

A: The ego vehicle.
\smallskip

Q: Would <c1,CAM\_BACK,514.2,503.3>/... take <c3,CAM\_FRONT,400.1,717.2>/... into account?

A: No.
\smallskip

Q: What object would consider <c1,CAM\_FRONT,985.8,516.7>/... to be most relevant to its decision?

A: The ego vehicle.
\smallskip

Q: Except for the ego vehicle, what object would consider <c1,CAM\_FRONT,985.8,516.7>/... to be most relevant to its decision?

A: <c2,CAM\_FRONT,1217.5,511.7>.
\smallskip

\textcolor{blue}{Signal Meaning}
\smallskip

Q: What does <c2,CAM\_BACK\_LEFT,400.8,654.2>/... mean?

A: No entry.
\smallskip

Q: What kind of traffic sign is <c2,CAM\_BACK\_LEFT,400.8,654.2>/...?

A: Traffic cone.
\smallskip

\textcolor{blue}{\textbf{Planning}}
\smallskip

\textcolor{blue}{Safe/Unsafe Action} 
\smallskip

Q: In this scenario, what are safe actions to take for the ego vehicle?

A: Decelerate gradually without braking, keep going at the same speed.
\smallskip

Q: In this scenario, what are dangerous actions to take for the ego vehicle?

A: Accelerate and go ahead, brake suddenly, drive backward, turn right.

\textcolor{blue}{Importance Ranking}
\smallskip

Q: What is the priority of the objects that the ego vehicle should consider? (in descending order)

A: <c2,CAM\_FRONT,514.7,462.2>, <c3,CAM\_FRONT,950.3,613.1>, <c1,CAM\_FRONT,707.5,472.5>.

\textcolor{blue}{Goal Action}
\smallskip

Q: What is the target action of the ego vehicle?

A: Go straight.
\smallskip

\textcolor{blue}{Planning and Reasoning}
\smallskip

Q: What actions could the ego vehicle take based on <c1,CAM\_FRONT,920.0,509.2>/...? Why take this action and what's the probability?

A: The action is to decelerate gradually without braking, the reason is to keep a safe distance, high.
\smallskip

Q: Based on <c3,CAM\_FRONT,1591.1,441.8>/... in this scene, what is the most possible action of the ego vehicle?

A: Decelerate gradually without braking.
\smallskip

\textcolor{blue}{Possible Collision}
\smallskip

Q: What is the probability of colliding with <c1,CAM\_FRONT,920.0,509.2>/... after the ego vehicle goes straight and keeps the same speed/accelerates and goes straight/...?

A: Low.
\smallskip

Q: What actions taken by the ego vehicle can lead to a collision with <c1,CAM\_FRONT,920.0,509.2>/...?

A: Accelerate and go straight.
\smallskip

\textcolor{blue}{Ego Attention} 
\smallskip

Q: What is the traffic signal that the ego vehicle should pay attention to?

A: None.
\smallskip

Q: Is <c1,CAM\_FRONT,920.0,509.2>/... an object that the ego vehicle should consider in the current scene?

A: Yes.
\smallskip

Q: Is it necessary for the ego vehicle to take <c3,CAM\_FRONT,400.1,717.2>/... into account?

A: Yes.

\textcolor{blue}{Free-form QA/Comment} 
\smallskip

Q: What impact does this situation have on driving vehicles?

A: The road scene is complex, please slow down.
\smallskip

Q: What's your comment on this scene?

A: Pedestrians at the intersection, please be careful and give way.
\smallskip

...

\end{tcolorbox}
\vspace{-8pt} \captionof{table}{\textbf{Question templates of \dataname-nuScenes at task level.} The categories in the table correspond to those in Fig.~\ref{fig:nus_statis}.}\label{box:nus_qa}
\vspace{0.15cm}
}
\smallskip
\bheading{Object level.} 
We also conduct some statistics at the object level since QAs in our \dataname-nuScenes revolve around key objects. 
\figref{fig:nus_obj_stats} (left) shows the distribution of our key object types.
Given the substantial differences in questions associated with traffic elements compared to other categories, we separately conduct statistics for QA types related to traffic elements and the remaining categories. 
The results are depicted in~\figref{fig:nus_obj_stats} (right).

\section{DriveLM-CARLA}
\label{supp:CARLA}

In this section, we introduce the details of \dataname-CARLA, including the dataset composition and collection methodology.

\subsection{Dataset Composition}
\label{supp:CARLA-data-comp}

\begin{figure}[t!]
    \centering
    \includegraphics[width=\linewidth]{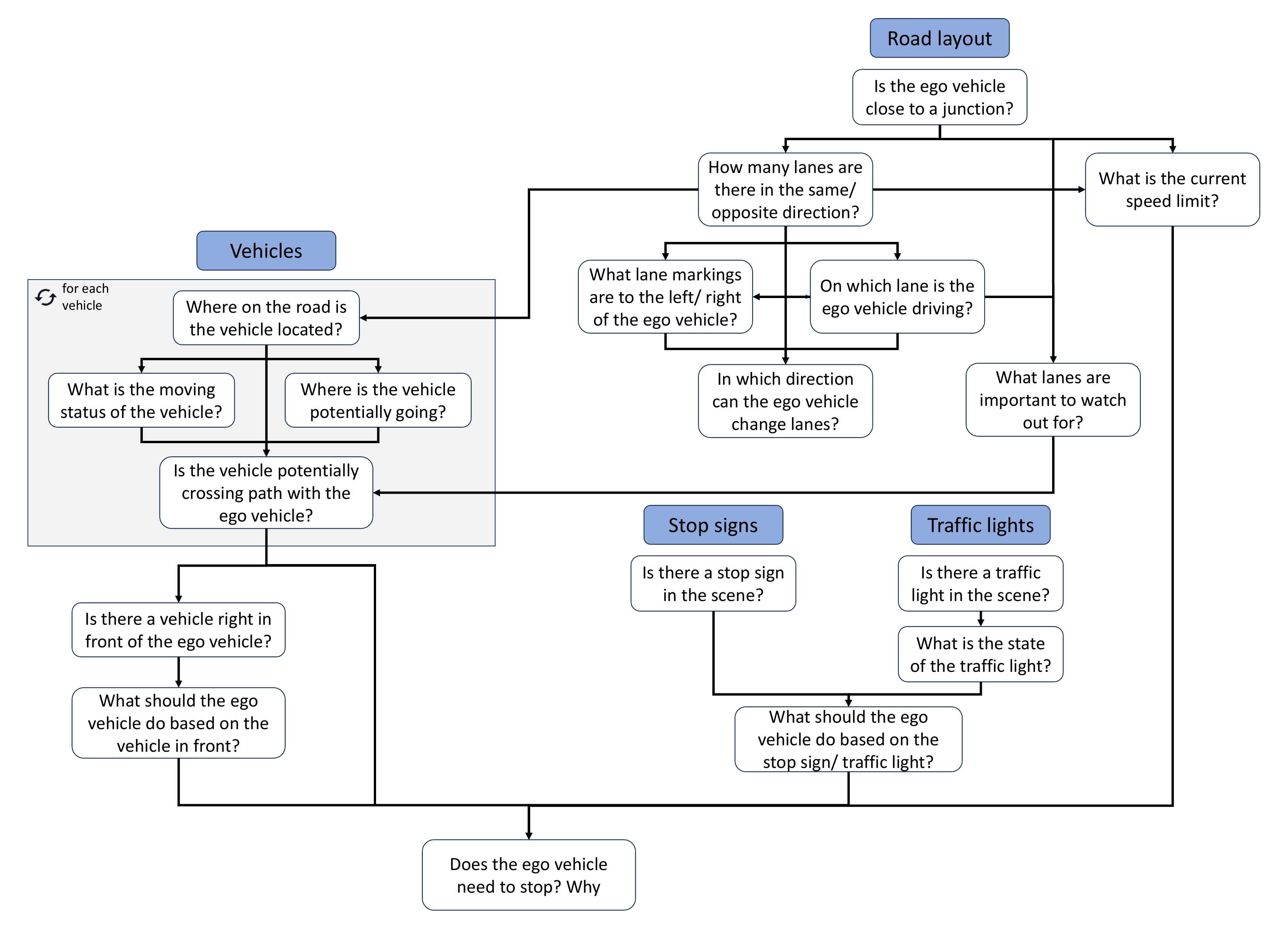}
    \caption{\textbf{Detailed flow of CARLA graph.} We show the full graph of \dataname-CARLA. The graph consists of questions and answers about the road layout, traffic lights, stop signs, and vehicles.}
    \label{fig: carla_graph_full}
\end{figure}

\dataname-CARLA consists of automatically generated frame-level question-answer pairs that are structured with an interconnected graph. The graph structure can be seen in Fig.~\ref{fig: carla_graph_full}. In the current version, the dataset consists of questions about the road layout, stop signs, traffic lights, and vehicles. In future versions, the dataset can be extended to more categories like static objects, weather, traffic signs, and others. 

Utilizing the driving simulator CARLA for the data generation process allows for scalable annotations and data without any manual effort involved. Additionally, the dataset supports a variety of sensor outputs from CARLA, including semantic segmentation, depth map, LiDAR, and others, which can be employed to train different network architectures.
Each question within the graph is designed in a way to facilitate situational reasoning, which could be instrumental in answering subsequent questions. As with \dataname-nuScenes, each question can be categorized into perception, prediction, or planning. 
For each QA-pair, besides the corresponding question and answer, we also save the object ID in case the QA-pair is about an object. This ID is consistent over time, enabling object tracking and temporal reasoning in future studies. In addition, relationships to parent and child questions within the graph are documented to allow efficient traversal of the graph.

\subsection{Expert Algorithm: PDM-Lite}
\label{supp:CARLA-expert}

\begin{figure}[bp]
    \centering
    \includegraphics[width=\linewidth]{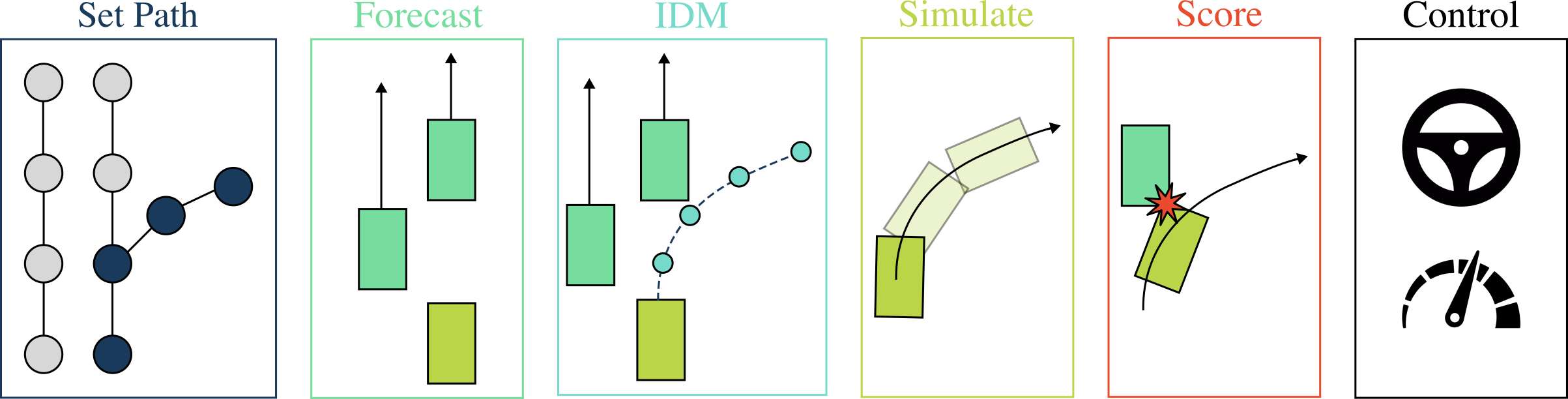}
    \caption{\textbf{PDM-Lite}. Our new planner which solves all 38 scenarios of CARLA Leaderboard 2.0 has 6 stages, detailed in \secref{supp:CARLA-expert}.}
    \label{fig:carla_pdm_lite}
\end{figure}

While previous CARLA expert algorithms like the privileged rule-based expert used by TransFuser++~\cite{Jaeger2023ICCV} can only solve the scenarios implemented in CARLA Leaderboard 1.0, PDM-Lite \cite{Beißwenger2024PdmLite} is designed to tackle all 38 scenarios present in Leaderboard 2.0. It consists of six distinct stages, summarized in \figref{fig:carla_pdm_lite}.

\boldparagraph{Path Planning} First, PDM-Lite creates a dense path in the form of spatially equidistant points using an A* planning algorithm, given sparse target points (up to 200m apart) by the leaderboard module. The plan is based on the HD map of the town, provided by CARLA. Furthermore, we add information such as speed limits and distances to the next traffic light/stop sign to different sections of this route that is to be traversed. To handle scenarios that require leaving the default path (\eg \textit{Accident, ConstructionObstacle, ParkedObstacle, VehicleOpensDoor, AccidentTwoWays, ConstructionObstacleTwoWays, ParkedObstacleTwoWays, VehicleOpensDoorTwoWays, YieldToEmergencyVehicle}), a short segment of the route where the scenario will be spawned is shifted laterally towards an adjacent lane. %

\boldparagraph{Agent Forecasting} PDM-Lite generates a forecast of dynamic agents for a horizon of 2s with a temporal resolution of 20Hz. Since we do not know other actors' paths and controls in advance, we assume they maintain their previous controls and apply similar ones in the near future. Using a kinematic bicycle model with parameters taken from~\cite{Chen2021ICCVa}, PDM-Lite forecasts other agents' motion using this constant action assumption, similar to the expert of~\cite{Jaeger2023ICCV}. We only consider actors closer than 50m to the ego agent for forecasting. %

\boldparagraph{IDM Target Speed} We generate a target speed proposal using the Intelligent Driver Model~\cite{Treiber2000}. Specifically, while the original IDM selects only one vehicle as a leading actor, we instead iteratively apply it to all vehicles, pedestrians, non-green traffic lights, and stop signs intersecting the path ahead of the ego vehicle. Once the leading actors are selected, we determine a target speed for each actor using the parameters summarized in \tabref{tab:pdm}. We use the minimum speed value obtained as the final target speed proposal for that timestep.

\begin{table}[t!]
    \vspace{-0.0cm}
    \caption{\textbf{Target Speed Proposal by IDM.} PDM-Lite uses IDM to select a target speed. The parameters for the desired net distance and desired time headway differ with respect to the type of the leading actor: vehicles, bicycles / stop signs / traffic lights / walkers / static objects (\eg construction signs).}
    \label{tab:pdm}
    \centering
        \arrayrulecolor{black} 
    \begin{tabular}{l|l|l}
        \toprule
    \textbf{Parameter} & \textbf{Value} & \textbf{Description}  \\ \midrule
    $v_0$ & $0.72$ $v_\textrm{lane}$ & Desired velocity. 72\% of the speed limit \\ 
    $s_0$ & $4.0$/$2.0$/$6.0$/$4.0$/$2.0$ m & Desired net distance to the leading agent \\ \
    $T$ & $0.25$/$0.1$/$0.1$/$0.1$/$0.1$ s & Desired time headway to leading agent\\ 
    $a$ & $24.0 \textrm{ ms}^{-2}$ & Maximum acceleration of ego \\ 
    $b_{v \leq 6.02}$ & $8.7 \textrm{ ms}^{-2}$ & Maximum deceleration of ego if $v \leq 6.02$\\ 
    $b_{v>6.02}$ & $3.72 \textrm{ ms}^{-2}$ & Maximum deceleration of ego if $v > 6.02$ \\ 
    $\delta$ & 4.0 & Acceleration exponent\\ 
    \bottomrule
    \end{tabular}
    \vspace{-0.0cm}
\end{table}

\boldparagraph{Simulation} We simulate the trajectory with the proposed target speed for 2s at 20Hz by alternatively applying the longitudinal controller (described in the following) and a kinematic bicycle model. Thereby, the proposal is converted into an actual expected sequence of ego-vehicle bounding boxes in closed-loop. %

\boldparagraph{Scoring} Having forecasted the bounding boxes for all actors, we can now check for bounding box intersections between the simulated ego vehicle and other vehicles. We score the ego vehicle's motion accordingly: if we detect an intersection with a non-leading and non-rear-end vehicle, we reject the IDM target speed proposal, and instead set the target speed to zero.
 
\boldparagraph{Controllers} Controlling the vehicle requires three values: steer, throttle, and brake. The steering value can be directly estimated from the ego vehicle's location, velocity, and path. We use a lateral PID controller for doing so, similar to~\cite{Jaeger2023ICCV}, which minimizes the angle to a selected point along the path ahead. For the throttle and brake predictions, we employ a linear regression model using features extracted based on the current speed and target speed.

\boldparagraph{Results} The performance of PDM-Lite is evaluated on the 20 official validation routes of Leaderboard 2.0. The results of three different seed evaluations are presented in \tabref{tab:pdm_results}. Routes 0-9 and 10-19 are identical except for weather conditions, providing an additional measure of performance variance. Since PDM-Lite utilizes privileged information, its variance in performance is not due to the differing weather parameters, but rather the random initialization of the surrounding traffic.
\begin{table}[htbp]
    \centering
    \caption{\textbf{PDM-Lite on CARLA Official Validation Routes}. We show the driving performance for each route, including Driving Score (DS), Infraction Score (IS), Route Completion (RC), Normalized Driving Score (NDS), and Normalized Infraction Score (NIS), as well as the average and their standard deviations for a total of 3 seeds. Routes 0-9 are identical to routes 10-19, except for the weather (which is ignored by PDM-Lite) and traffic behaviors.}
    \label{tab:pdm_results}
    \begin{tabular}{c|c|c|c|c|c}
        \toprule
        \textbf{Route} & \textbf{DS} & \textbf{RC} & \textbf{IS} & \textbf{NDS} & \textbf{NIS}\\
        \midrule
        0 & 14.0 {\color{gray}\small $\pm 0.0$} & 100.0 {\color{gray}\small $\pm 0.0$} & 0.14 {\color{gray}\small $\pm 0.00$} & 48.2 {\color{gray}\small $\pm 0.0$} & 0.48 {\color{gray}\small $\pm 0.00$}\\
        1 & 56.2 {\color{gray}\small $\pm 37.4$} & 71.1 {\color{gray}\small $\pm 40.9$} & 0.75 {\color{gray}\small $\pm 0.18$} & 61.9 {\color{gray}\small $\pm 42.1$} & 0.69 {\color{gray}\small $\pm 0.32$}\\
        2 & 55.3 {\color{gray}\small $\pm 14.3$} & 100.0 {\color{gray}\small $\pm 0.0$} & 0.55 {\color{gray}\small $\pm 0.14$} & 79.3 {\color{gray}\small $\pm 6.8$} & 0.79 {\color{gray}\small $\pm 0.07$}\\
        3 & 18.4 {\color{gray}\small $\pm 7.8$} & 74.7 {\color{gray}\small $\pm 35.8$} & 0.34 {\color{gray}\small $\pm 0.20$} & 38.9 {\color{gray}\small $\pm 20.4$} & 0.51 {\color{gray}\small $\pm 0.07$}\\
        4 & 59.8 {\color{gray}\small $\pm 13.5$} & 100.0 {\color{gray}\small $\pm 0.0$} & 0.60 {\color{gray}\small $\pm 0.14$} & 80.5 {\color{gray}\small $\pm 6.8$} & 0.81 {\color{gray}\small $\pm 0.07$}\\
        5 & 34.9 {\color{gray}\small $\pm 9.0$} & 68.0 {\color{gray}\small $\pm 0.0$} & 0.51 {\color{gray}\small $\pm 0.13$} & 46.2 {\color{gray}\small $\pm 5.6$} & 0.68 {\color{gray}\small $\pm 0.08$}\\
        6 & 36.4 {\color{gray}\small $\pm 7.9$} & 100.0 {\color{gray}\small $\pm 0.0$} & 0.36 {\color{gray}\small $\pm 0.08$} & 70.3 {\color{gray}\small $\pm 5.1$} & 0.70 {\color{gray}\small $\pm 0.05$}\\
        7 & 3.5 {\color{gray}\small $\pm 1.8$} & 100.0 {\color{gray}\small $\pm 0.0$} & 0.03 {\color{gray}\small $\pm 0.02$} & 25.7 {\color{gray}\small $\pm 5.2$} & 0.26 {\color{gray}\small $\pm 0.05$}\\
        8 & 15.3 {\color{gray}\small $\pm 3.3$} & 100.0 {\color{gray}\small $\pm 0.0$} & 0.15 {\color{gray}\small $\pm 0.03$} & 52.1 {\color{gray}\small $\pm 3.7$} & 0.52 {\color{gray}\small $\pm 0.04$}\\
        9 & 73.3 {\color{gray}\small $\pm 18.9$} & 100.0 {\color{gray}\small $\pm 0.0$} & 0.73 {\color{gray}\small $\pm 0.19$} & 89.8 {\color{gray}\small $\pm 7.2$} & 0.90 {\color{gray}\small $\pm 0.07$}\\
        \midrule
        10 & 12.8 {\color{gray}\small $\pm 7.6$} & 100.0 {\color{gray}\small $\pm 0.0$} & 0.13 {\color{gray}\small $\pm 0.08$} & 44.0 {\color{gray}\small $\pm 10.2$} & 0.44 {\color{gray}\small $\pm 0.10$}\\
        11 & 60.5 {\color{gray}\small $\pm 32.0$} & 100.0 {\color{gray}\small $\pm 0.0$} & 0.61 {\color{gray}\small $\pm 0.32$} & 79.5 {\color{gray}\small $\pm 18.1$} & 0.79 {\color{gray}\small $\pm 0.18$}\\
        12 & 52.0 {\color{gray}\small $\pm 11.3$} & 100.0 {\color{gray}\small $\pm 0.0$} & 0.52 {\color{gray}\small $\pm 0.11$} & 78.9 {\color{gray}\small $\pm 6.5$} & 0.79 {\color{gray}\small $\pm 0.06$}\\
        13 & 26.3 {\color{gray}\small $\pm 15.9$} & 49.4 {\color{gray}\small $\pm 35.8$} & 0.58 {\color{gray}\small $\pm 0.07$} & 32.2 {\color{gray}\small $\pm 28.5$} & 0.57 {\color{gray}\small $\pm 0.11$}\\
        14 & 39.4 {\color{gray}\small $\pm 20.8$} & 100.0 {\color{gray}\small $\pm 0.0$} & 0.39 {\color{gray}\small $\pm 0.21$} & 69.0 {\color{gray}\small $\pm 11.0$} & 0.69 {\color{gray}\small $\pm 0.11$}\\
        15 & 24.0 {\color{gray}\small $\pm 17.2$} & 68.0 {\color{gray}\small $\pm 0.0$} & 0.35 {\color{gray}\small $\pm 0.25$} & 35.8 {\color{gray}\small $\pm 13.8$} & 0.53 {\color{gray}\small $\pm 0.20$}\\
        16 & 42.0 {\color{gray}\small $\pm 0.0$} & 100.0 {\color{gray}\small $\pm 0.0$} & 0.42 {\color{gray}\small $\pm 0.00$} & 73.9 {\color{gray}\small $\pm 0.0$} & 0.74 {\color{gray}\small $\pm 0.00$}\\
        17 & 4.1 {\color{gray}\small $\pm 1.2$} & 95.3 {\color{gray}\small $\pm 6.7$} & 0.04 {\color{gray}\small $\pm 0.01$} & 25.6 {\color{gray}\small $\pm 5.5$} & 0.27 {\color{gray}\small $\pm 0.04$}\\
        18 & 12.1 {\color{gray}\small $\pm 7.9$} & 100.0 {\color{gray}\small $\pm 0.0$} & 0.12 {\color{gray}\small $\pm 0.08$} & 43.7 {\color{gray}\small $\pm 15.6$} & 0.44 {\color{gray}\small $\pm 0.16$}\\
        19 & 86.7 {\color{gray}\small $\pm 18.9$} & 100.0 {\color{gray}\small $\pm 0.0$} & 0.87 {\color{gray}\small $\pm 0.19$} & 94.9 {\color{gray}\small $\pm 7.2$} & 0.95 {\color{gray}\small $\pm 0.07$}\\
        \midrule
        \textbf{Avg.} & \textbf{36.3} {\color{gray}\small $\pm 28.1$} & \textbf{91.3} {\color{gray}\small $\pm 21.1$} & \textbf{0.41} {\color{gray}\small $\pm 0.28$} & \textbf{58.5} {\color{gray}\small $\pm 25.9$} & \textbf{0.63} {\color{gray}\small $\pm 0.22$}\\
        \bottomrule
    \end{tabular}
\end{table}

\subsection{Collection Methodology}
\label{supp:CARLA-coll-method}
In this section, we provide details about the data collection and the annotation process.

\bheading{Simulator settings.} 
We utilize the CARLA Simulator (version 0.9.14) with Leaderboard 2.0~\cite{Dosovitskiy17} to generate our dataset. Leaderboard 2.0 introduces two new large maps along with a suite of new scenarios, enhancing the diversity of the training and evaluation environments. Town 12 serves as the training town, while Town 13 is reserved for evaluation. Each town covers an area of 10 x 10 square kilometers, encompassing varied environments such as rural, residential, and urban landscapes to replicate real-world driving conditions. The CARLA team provides a total of 90 training routes, spanning 780.6 kilometers, and 20 evaluation routes, measuring 247.6 kilometers. Every route incorporates multiple driving scenarios. We segmented these routes into shorter segments, approximately 150 meters in length and filter routes that start and end at the same position. The traffic manager within Leaderboard 2.0 initializes random background traffic around the ego vehicle comprised exclusively of 'car' entities. To enrich the dataset with greater diversity, we introduce additional vehicle classes including 'trucks', 'vans', 'bicycles', and 'motorcycles'. Moreover, we implemented randomized weather configurations for each training and evaluation route to mimic realistic driving conditions. However, night-time settings were excluded from our study due to the inadequate illumination in certain map regions. Low-light conditions significantly impede the correctness of the automatic labeling process since it is hard to obtain information about the visibility of certain objects in the image.

\smallskip
\bheading{Data collection.} 
We execute the expert on each of the routes and gather a comprehensive set of sensor data.
The sensor data includes: (1) \textit{RGB image}, (2) \textit{LiDAR point cloud}, (3) \textit{semantic segmentation images}, (4) \textit{depth maps}, (5) \textit{Bird's Eye View (BEV) semantic segmentation}. While \algname{} leverages only RGB images, retraining TransFuser++ needs the additional data for the auxiliary tasks. In addition, we extract privileged information from the simulator about the status of the static and dynamic objects in the scene, as follows:
\begin{itemize}
    \item \textit{Ego vehicle:} 3D bounding box, speed, brake, id
    \item \textit{Other vehicles:} 3D bounding box, number of lidar points inside BB, distance to ego, speed, steer, throttle, brake, id, color, vehicle type, number of wheels, traffic light state, lane information (i.e., on which road and lane is the vehicle driving), vehicle in junction or not, distance to next junction, next high-level command
    \item \textit{Pedestrians:} 3D bounding box, number of lidar points inside BB, gender, age, distance to ego, speed, id, lane information
    \item \textit{Traffic lights:} 3D bounding box, distance to ego, state, affects ego vehicle
    \item \textit{Stop signs:} 3D bounding box, distance to ego, affects ego vehicle
    \item \textit{Static cars (parked cars):} 3D bounding box, lane information
    \item \textit{Landmarks (e.g., speed signs):} 3D bounding box, distance to ego, id, text, value
    \item \textit{Weather:} weather parameters
\end{itemize}

\smallskip
\bheading{Language labels.} Based on the information we extract from the simulator we create questions and answers with hand-crafted sentence templates. For more linguistic diversity and to prevent overfitting to those sentence structures those sentences could be further augmented with current state-of-the-art language models like GPT-4. However, in this work, we use a version of the dataset that is not augmented.

\section{DriveLM-Metrics}
\label{supp:metrics}

In this section, we offer a detailed introduction to DriveLM-Metrics. 
DriveLM-Metrics can be broadly categorized into three parts: $P_{1-3}$ VQA Metrics, Behavior Task Metrics, and Motion Task Metrics.

\subsection{$P_{1-3}$ VQA Metrics}
\label{supp:metrics-vqa}

We assess the performance of $P_{1-3}$ using common VQA metrics, and we introduce the \textit{GPT score} for a more semantically comprehensive evaluation of our QA results. 
Additionally, given the graph structure of our QA, we propose the \textit{Completeness} score to provide a thorough assessment.

\bheading{BLEU}~\cite{papineni2002bleu} measures the similarity between a generated text and one or more reference texts. 
It operates by comparing n-grams in the generated text to those in the reference texts, with higher precision indicating a better match. 
However, the BLEU score exhibits insensitivity to semantic nuances and variations in word order.

\bheading{ROUGE\_L}~\cite{lin2004rouge} calculates scores with the longest common sub-sequence of the model outputs and the reference answers. 
Similar to the BLEU metric, ROUGE is used to assess the level of matching between generated results and standard references, with the key difference being that ROUGE is based on recall. 

\bheading{METEOR}~\cite{lavie-agarwal-2007-meteor} takes into account precision, recall, stemming, synonymy, stemming, and word order. 
It establishes alignment between model outputs and references, computes the 1-gram matching between them, and then applies penalties based on chunk blocks, providing a more nuanced evaluation.

\bheading{CIDEr}~\cite{vedantam2015cider} combines elements from BLEU and vector space models. 
The underlying concept involves treating each sentence as a document, calculating its n-gram TF-IDF vector, and using cosine similarity to measure the semantic consistency between candidate and reference sentences.
CIDEr captures matches between n-grams of different lengths and differentiates the importance of various n-grams through TF-IDF weighting.

\bheading{SPICE}~\cite{anderson2016spice} first parses the text into a syntactic dependency tree using Probabilistic Context-Free Grammar~\cite{jelinek1992basic}, then maps the dependency tree into a scene graph in a rule-based manner.
The scene graph describes the objects, attributes, and their relationship in the original text, and the SPICE score is computed as the F-score of the generated scene graphs from prediction and ground truth.

\bheading{GPT Score} is a metric provided by ChatGPT. Traditional metrics mainly assess word-level performance and may not capture semantic nuances, potentially yielding unexpected evaluation outcomes. Leveraging ChatGPT's robust reasoning capabilities, we employ it to gauge prediction quality and derive a more rational score. ChatGPT is prompted to assign a numerical score between 0 and 100, with higher scores indicative of enhanced prediction accuracy. The detailed prompt for GPT score evaluation is shown in Table~\ref{box:gpt_score}.

{
\captionsetup{type=table}
\begin{tcolorbox}[colback=gray!10,
                  colframe=black,
                  width=\linewidth,
                  arc=1mm, auto outer arc,
                  boxrule=0.5pt,
                 ]

{\cmtt{\textcolor{blue}{\textbf{Messages}}
 = [ 

\smallskip
 \{
"role": "system",
"content": f"""} An evaluator who rates my answer based on the correct answer.
\cmtt{""" \},}

\smallskip
\cmtt{\{
"role": "user",
"content": f"""} Rate my answer based on the correct answer out of 100, with higher scores indicating that the answer is closer to the correct answer, and you should be accurate to single digits like 62, 78, 41, etc. This is the correct answer: 
\{\cmtt{\textcolor{blue}{GT}\}}. This is my answer: \{\cmtt{\textcolor{blue}{Pred}\}}.
"""\}]}

\end{tcolorbox}
\vspace{-8pt} 
\captionof{table}{\textbf{Prompt for GPT score.} This differs from the prompt used in DriveGPT4~\cite{xu2023drivegpt4}, but the resulting score is similar.}\label{box:gpt_score}
\vspace{0.15cm}
}

\bheading{Completeness} provides a score that accounts for how many ground truth questions are correctly answered associated with a frame.
For each QA, if the score of the predicted answer is above a threshold, then this QA is considered ``correctly answered'' and is a correct prediction, otherwise an incorrect prediction.
We then calculate the accuracy, which is the ratio of the number of correct predictions to the total number of predictions.
In our setting, we utilize the SPICE score and set the threshold at 0.5.

\subsection{Behavior Task Metrics}
\label{supp:metrics-behavior}
We evaluate behavior predictions by classification accuracy, along with a breakdown of the overall accuracy into its steering and speed components.

\bheading{Classification Accuracy} is the metric we use to evaluate Behavior Prediction Task, comprising \emph{accuracy of behavior, behavior speed, and behavior steer}. 
Specifically, the ground truth of the ego vehicle future trajectory is a set of $N$ points with coordinates $(x,y)$ under the bird's-eye-view, noted as $\{(x_0,y_0), (x_1,y_1) , ..., (x_N, y_N)\}$.
Each point denotes the offset of the future position to the current position by a number of fixed interval times.
Then, the distance for $x,y$ at each time interval is independently computed as: 
\begin{equation}
{\{x,y\}}_{dist} = (({\{x,y\}}_1 - {\{x,y\}}_0), ..., ({\{x,y\}}_N - {\{x,y\}}_{N-1}))
\end{equation}\label{eq:xy_dist_supp}
The mean of $x_{dist}$ and $y_{dist}$ are mapped to one of the predefined bins, where each bin corresponds to a category in either speed or steering, noted as $B_{speed}$ and $B_{steer}$ respectively.
Finally, the speed and steering categories for this trajectory form the {behavior} category as $(B_{speed}, B_{steer})$.
We compare them with the behaviors of our \algname outputs and calculate the related accuracy.

\subsection{Motion Task Metrics}
For measuring the performance of the motion stage, we use standard metrics from the nuScenes and Waymo benchmarks: average and final displacement error, (ADE, FDE), and the collision rate of the predicted trajectory.

\bheading{ADE} stands for Average Displacement Error, indicating the average L2 distance between the predicted trajectory and the ground truth trajectory over all predicted horizons.
It is the average of the errors at 1$^{st}$, 2$^{nd}$ and 3$^{rd}$ second.

\bheading{FDE} stands for Final Displacement Error, which measures the Euclidean distance between the predicted endpoint and the true endpoint at the last predicted step (the 3$^{rd}$ second).

\bheading{Collision Rate} accounts for the ratio of how many test frames the predicted trajectory collides with objects in over all test frames.
The number reported in Table~\textcolor{red}{2} of the main paper is the average of the collision rate at 1$^{st}$, 2$^{nd}$ and 3$^{rd}$ second.

Note that the calculation of \textbf{ADE}, \textbf{FDE} and \textbf{Collision Rate} follows the setting used in UniAD~\cite{hu2023_uniad} but not ST-P3~\cite{hu2022stp3}.
For example, in terms of the FDE and the collision rate at 3$^{rd}$ second, the UniAD setting will consider the error/collision rate at only this timestep, while the ST-P3 setting will consider the error/collision rate as an average over 0.5, 1, 1.5, 2, 2.5, 3 seconds.
For more details, please refer to the UniAD repo \href{https://github.com/OpenDriveLab/UniAD/commit/ffac1a69a6d01c9dca5cd2ed751007896df351f8}{discussion}.
Additionally, please note that errors reported on the full nuScenes validation dataset (in prior work) is not directly comparable to results reported on the \dataname-nuScenes val split, a challenging subset of this consisting of only keyframes with intention changes.

\section{DriveLM-Agent}
\label{supp:agent}

In this section, we introduce the details of \algname, including the graph prompting scheme and the trajectory tokenization process.

\subsection{Prompting with Context}

In terms of the implementation, the content of context differs during the training and inference of \algname, following the teacher-forcing setting~\cite{lamb2016professor,toomarian1992learning} generally adopted in recurrent networks. During training, for each edge $e\!\in\!E$ in the frame, we pick the child QA. The child questions in the edges are appended with the ground truth parent QA as the context.
All QA pairs are used during training, including those without context.
The objective used is next token prediction, the standard approach for language modeling.
During inference, the model is applied interactively in multiple rounds to get the required context predictions as inputs for each child question.
Specifically, the model is prompted with the five stages of questions in the sequential order of $P_1, P_2, P_3, B, M$.
In this order, the model can only infer the questions in the succeeding stages after getting the predicted answer from the preceding stages.

\subsection{Trajectory Tokenization Details}

To generate action sequences (\textit{i.e.}, ego future trajectories) directly with the language model we use for building the graph, we adopt the method of RT-2~\cite{zitkovich2023rt}. This process entails the discretization and tokenization of the continuous trajectory. 

Initially, we analyze the distribution of the future trajectories within the nuScenes dataset. To effectively convert the continuous \textit{(x, y)} coordinate space into a discrete set of actions, we partition each coordinate axis into 256 discrete intervals. This granularity ensures a sufficient level of detail while maintaining a manageable number of tokens for the language model.

Each discretized bin corresponds to a unique token within the vocabulary of the language model. We extract the token identifiers (IDs) for numeric tokens within the vocabulary. To ensure coherence and preserve the ability to express numerical values, we omit single-digit tokens from this mapping process. Out of the remaining numeric tokens, a subset of 256 token IDs is selected to represent the trajectory data. In addition to these, we introduce two special tokens designated for marking the start and end of a trajectory sequence – the start-of-trajectory (SOT) token and the end-of-trajectory (EOT) token, respectively.
This tokenization scheme enables us to encode complex trajectory information as a sequence of tokens that a language model can process. Using this mapped vocabulary, the language model can generate predicted future trajectory sequences by outputting a series of tokens, which are then translated back into the coordinate space.

\begin{figure*}[t!]
    \centering
    \includegraphics[width=\linewidth]{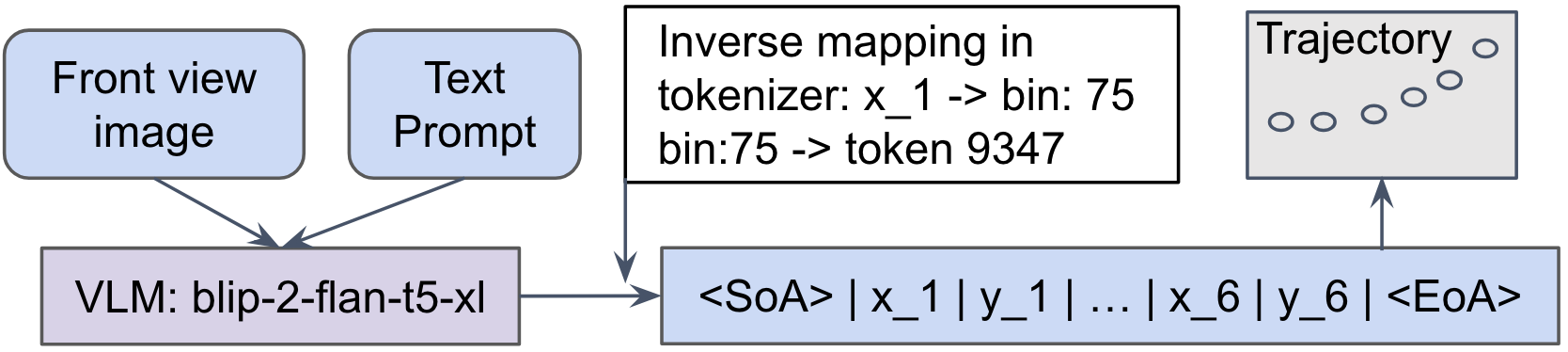}
    \caption{\textbf{Detailed architecture of DriveLM-Agent.} An inverse mapping is designed to embed the real-world ego-vehicle trajectory into the token space of blip-2-flan-t5-xl.}
    \label{fig: simple_pipeline}
\end{figure*}

\section{Experiments}
\label{supp:exp}

In this section, we introduce the details of experiments, including implementation details of each subsection in \textbf{Section \textcolor{red}{4}} in the main paper, more metrics on the VQA part, more ablation and comparison on computational complexity.

\subsection{Implementation Details}

Here we explain the implementation details for the training and validation settings used in our experiments of each of each subsections in \textbf{Section~\textcolor{red}{4}} in the main paper.

\bheading{Fine-tuning Details.}
We configure the learning rate as 0.0001, no learning-rate scheduler, random seed as 1234, and other settings following the default LoRA~\cite{hu2021lora} configuration.
For the BLIP-2 model, we use a maximal sequence length of 400, and other hyperparameters remain the same as the official BLIP-2 implementation.

\bheading{Implementation Details for Experiment in Section~\textcolor{red}{4.1} \&~\textcolor{red}{4.3} \&~\textcolor{red}{4.4}.}

During training, we utilize all QAs as input per frame, with a subset of them having contexts (questions from $P_{2,3}$, $B$, and $M$).
The contexts are extracted from ground truth, following the teacher-forcing setting~\cite{lamb2016professor,toomarian1992learning} generally adopted in recurrent networks.  
As for inference, due to the variant complexity of the scenarios, the count of $P_{1-3}$ QA per frame is highly imbalanced across the dataset, with a variance of over 260 on \dataname-nuScenes.
To balance the impact of this, 
we compute the GVQA Scores on only a subset of QA associated with each frame.
To extract the QA subset, we design a set of QA patterns for each stage based on the questions generally associated with that stage. 
We ensure that for all our validation frames, each stage has at least one question matched with the designed pattern. 
In this process, except for the questions in stage $P_1$, all questions in other stages have context from the previous stage's QA, where the answers are derived from the prediction in the preceding steps.
Two specific graph structure examples can be found in~\figref{fig:nus_detailed_results} and~\figref{fig:nus_results}.

\bheading{Implementation Details for Experiment in Section~\textcolor{red}{4.2}.}
The model is trained in the same scheme as in Section~\textcolor{red}{4.1} \&~\textcolor{red}{4.3} \&~\textcolor{red}{4.4}, and the training set is the \dataname-nuScenes train split.
During the inference, as there are no annotation on the Waymo (not even questions), we devise the question in a rule-based manner.
Specifically, we re-use the general-purpose questions in the perception stage from \dataname-nuScenes for Waymo as the starting questions.
Then we try to find if there is any objects in the answer that is matched in the \dataname-nuScenes annotation, such as ``pedestrians'', ``cars'', ``trucks'' and so on.
Then, we generate the questions based on those matched objects automatically, which serve as the following questions in the prediction and the planning stages.
A specific graph structure example can be found in~\figref{fig:waymo_results}.

\bheading{Implementation Details for Experiment in Section~\ref{sec:exp_carla}.}
For the generalization experiment in Section~\ref{sec:exp_carla} we add two new questions to the graph: (1) \textit{Is there a person in the scene?} and (2) if the answer to the first question is yes we ask \textit{What should the vehicle do based on the pedestrian that is crossing the road?} if the answer is no we ask \textit{What should the ego vehicle do?}.
The answer to the second question gets concatenated to the context of the final behavior question. 
Three examples can be found in~\figref{fig:carla_resultsPed}.

\subsection{Generalization to Unseen Objects}
\label{sec:exp_carla}

Next, we evaluate zero-shot generalization to novel objects. \dataname-CARLA is collected without any pedestrians in the training or validation splits. We now generate a new test set called \dataname-CARLA-ped, which only consists of frames where a pedestrian is present in the scene. The correct behavior is to stop for the pedestrian.

\smallskip
\boldparagraph{Baselines} For this experiment, we compare \algname to TransFuser++~\cite{Jaeger2023ICCV}, the state-of-the-art for CARLA. It uses a larger input image, an additional LiDAR sensor, and several driving-specific annotations (depth, semantics, 3D bounding boxes, HD map) in comparison to \algname. However, because of these task-specific inputs and outputs, TransFuser++ can only be trained on the base \dataname-CARLA dataset and cannot incorporate general computer vision data during training, which makes generalization more challenging.

\smallskip
\boldparagraph{\algname} Taking advantage of the more general architecture of a VLM, we include samples from COCO~\cite{lin2014microsoft} and GQA~\cite{hudson2019gqa} along with \dataname-CARLA during training for \algname. We compare several versions: (1) we investigate the addition of a new $P_1$ question during inference, \textit{`Is there a person crossing the road?'} (`+ Pedestrian QA'). (2) As an upper bound, we directly input the ground truth $P_{1-3}$ graph to the model during inference, instead of the model's predictions. For more details, please refer to the supplementary material.

\begin{table}[t!]
\caption{\textbf{Generalization in \dataname-CARLA.} All methods underperform on \dataname-CARLA-ped with the novel pedestrian object, but \algname can be significantly improved by including a pedestrian-specific question in its GVQA graph. {move this to supp??}}
\centering
\resizebox{0.8\textwidth}{!}{
\begin{tabular}{l|rrr|rrr}
\toprule
    \multirow{2}{*}{\textbf{Method}}& \multicolumn{3}{r|}{\textbf{\dataname-CARLA} ($B$)} & \multicolumn{3}{r}{\textbf{\dataname-CARLA-ped} ($B$)}\\
    \cmidrule{2-7}
     & Acc. $\uparrow$ & Spd. $\uparrow$ & Str. $\uparrow$ & Acc. $\uparrow$ & Spd. $\uparrow$ & Str. $\uparrow$ \\
    \midrule
    TransFuser++~\cite{Jaeger2023ICCV} & \textbf{70.19} & \textbf{73.29} & \textbf{90.68} & 8.72 & 8.72 & 100.00\\
    \midrule
    \rowcolor[gray]{0.9}\algname &  59.63 & 61.50&78.26 & 4.59 & 4.59 & 100.00 \\
    \rowcolor[gray]{0.9} + Pedestrian QA & 52.17 & 55.28 & 77.64 &  \textbf{27.04} & \textbf{27.04} & 100.00 \\
    \midrule
    \textit{\algname (GT)} & \textit{60.25} & \textit{65.22} & \textit{80.12} & \textit{20.92} & \textit{20.92} & \textit{100.00} \\
    \textit{+ Pedestrian QA} & \textit{60.25} & \textit{65.22} & \textit{80.12} & \textit{92.35} & \textit{92.35} & \textit{100.00}\\
\bottomrule
\end{tabular}}
\label{tab:carla}
\end{table}

\boldparagraph{Results} We present our findings in \tabref{tab:carla}. We observe that TransFuser++ struggles on \dataname-CARLA-ped relative to \dataname-CARLA, with a drop in accuracy from $70.19$ to just $8.72$. \algname experiences a similar drop from $59.63$ to $4.59$. However, adding the pedestrian QA significantly boosts performance on the generalization setting to $27.04$, albeit with slightly reduced accuracy on regular scenes. This is mainly attributed to the fact that the VLM is not able to detect all pedestrians correctly. This indicates that the large performance gains of recently published VLMs~\cite{gao2023llamaadapterv2,peng2023kosmos} can support even better generalization ability in the domain of driving.  Additionally, when the pedestrian QA is provided in the privileged setting that assumes access to perfect context for each question in the graph, \algname achieves a near-perfect score ($20.92\rightarrow92.35$) on the frames with pedestrians
Note that \dataname-CARLA-ped only contains pedestrians crossing a straight road, so all models obtain a 100\% accuracy on the steering class (which is always \texttt{straight}).

\subsection{Results with More Metrics in VQA}
In \tabref{tab:vqa_more_metrics}, we provide the performance under BLEU-4~\cite{papineni2002bleu}, METEOR~\cite{lavie-agarwal-2007-meteor}, CIDEr~\cite{vedantam2015cider} and ROUGE-L~\cite{lin2004rouge} of the Table~\textcolor{red}{4} in the main paper.
One key observation is that different metrics reflect different trends in the performance, and the improvement is not consistent across all these metrics.
This brings us the motivation to use the GPT Score as the main metric in the VQA evaluation part.

\begin{table*}[t!]
\caption{\textbf{Graph-structured reasoning facilitates improved VQA with VLMs.} Completeness measures how many questions are correctly answered in one frame of data.
The improvement trends are not consisten across different conventional metrics, thus we need the GPT score as the main metrics as it evaluates the performance more comprehensively.
}
\centering
\resizebox{\linewidth}{!}
{
\begin{tabular}{l|ccccccc|ccccccc}
\toprule
\multirow{3}{*}{\textbf{Context}} & \multicolumn{14}{c}{\textbf{\dataname-nuScenes}}  \\
\cmidrule{2-15}
& \multicolumn{7}{c|}{Off-the-Shelf BLIP-2} & \multicolumn{7}{c}{\algname}  \\
\cmidrule{2-15}
& BLEU-4 $\uparrow$ & METEOR $\uparrow$ & CIDEr $\uparrow$ & ROUGE\_L $\uparrow$ & SPICE $\uparrow$ & GPT $\uparrow$ & Comp. $\uparrow$ & BLEU-4 $\uparrow$ & METEOR $\uparrow$ & CIDEr $\uparrow$ & ROUGE\_L $\uparrow$ & SPICE $\uparrow$ & GPT $\uparrow$ & Comp. $\uparrow$  \\
\midrule
None & 0.022 & 3.317 & 0.1185 & 7.205 & 4.336 & 42.97 & 1.064 & 51.89 & 35.81 & 2.463 & 66.79 & 42.56 & 71.39 & 30.04  \\
Graph & 0.022 & 3.882 & 0.0771 & 7.397 & 7.710 & 45.21 & 0.859 & 53.09 & 36.19 & 2.786 & 65.58 & 49.54 & 72.51 & 31.66  \\
\midrule
\textit{GT} & \textit{0.022} & \textit{4.397} & \textit{0.0758} & \textit{8.033} & \textit{8.192}  & \textit{41.10} & \textit{1.315} & \textit{53.06} & \textit{36.64} & \textit{3.069} & \textit{66.69} & \textit{50.29} & \textit{72.94} & \textit{32.41}  \\
\bottomrule
\end{tabular}
}
\vspace{-0.3cm}
\label{tab:vqa_more_metrics}
\end{table*}

\subsection{Stage-wise Ablation on Zero-shot Generalization across Sensor Configurations}
In \tabref{tab:waymo_supp}, we provide more settings of context in the zero-shot generalization across sensor configurations as in Table~\textcolor{red}{2} in the main paper.
One key observation is that the higher the accuracy of the behavior task, the better the performance of the motion task.
With more context in the behavior task, the accuracy improvement mainly originates from the improvement of the speeding accuracy, which finally affects the FDE score.

\subsection{More VLMs evaluated on \dataname-nuScenes}

\begin{table}[h!]
\vspace{-0.3cm}
\caption{\textbf{More VLMs evaluation on \dataname-nuScenes}. On the base model of LLaMA-Adapter V2, we observe a slight improvement by using multi-frame as input.}
    \centering
    \resizebox{0.8\textwidth}{!}{
    \begin{tabular}{c|cc|ccc|cc}
\toprule
    {\textbf{Table \textcolor{red}{A}.}} & {\textbf{Behavior}} & {\textbf{Motion}} & \multicolumn{3}{c|}{\textbf{Behavior ($B$)}} & \multicolumn{2}{c}{\textbf{Motion ($M$)}} \\
    \textbf{Base VLM} & \textbf{Context} & \textbf{Context} & Acc. $\uparrow$ & Speed $\uparrow$  & Steer $\uparrow$ & ADE $\downarrow$ & Col. $\downarrow$  \\
    \midrule
    {blip-2-flan-t5-xl} & Graph & $B$ & 57.49 & \textbf{69.89} & 80.63 & \textbf{1.74} & 1.89 \\
    \rowcolor[gray]{0.9} LLaMA-Adapter V2 & Graph & $B$ & 55.31 & 61.97 & 81.43 & 1.86 & 2.03 \\
    \rowcolor[gray]{0.9} LLaMA-Adapter V2$^{*}$ & Graph & $B$ & \textbf{57.83} & 67.21 & \textbf{83.49} & 1.75 & \textbf{1.69} \\
\bottomrule
\end{tabular}
}
\vspace{-0.4cm}
\label{tab:llama}
\end{table}

In terms of validating the generality of the proposed \algname and exploring if the video input helps the model's performance in this task, we select LLaMA-Adapter V2 (7B)~[\textcolor{red}{I}] as it is compatible with single- and multi-frame inputs. The \colorbox{lgray}{results} in \tabref{tab:llama} (within 9 training epochs) show performance similar to BLIP-2.

Note that due to the large variety of question types in DriveLM-nuScenes, analyzing each question would be prohibitively expensive.

Instead, we provided an ablation of the effectiveness of each stage ($P_{1-3}$) in \tabref{tab:waymo_supp}, and a representative-question-wise analysis.

We have to admit that it is non-trivial in improving the QA performance (P1-3) under current question setting and we found that it had little affect on the final task performance (behavior and motion).

This inspired us that maybe the question setup worth more effort to explore, and with well-set question, the model could be ``prompted correctly'' to understand the scene and perform the downstream task better.

\begin{table}[t!]
\caption{\textbf{Zero-shot Generalization across Sensor Configurations.} $B$* denotes using ground truth behavior QA as context for motion task. Results on 1k randomly sampled frames from the Waymo \texttt{val} set after training on \dataname-nuScenes.
Our key observation is that the higher the accuracy of the behavior task, the better the performance of the motion task.
}
\centering
\resizebox{0.8\linewidth}{!}{
\begin{tabular}{l|cc|ccc|cc}
\toprule
    \multirow{2}{*}{\textbf{Method}} & {\textbf{Behavior}} & {\textbf{Motion}} & \multicolumn{3}{c|}{\textbf{Behavior ($B$)}} & \multicolumn{2}{c}{\textbf{Motion ($M$)}} \\
    & \textbf{Context} & \textbf{Context} & Acc. $\uparrow$ & Speed $\uparrow$  & Steer $\uparrow$ & ADE $\downarrow$ & FDE $\downarrow$  \\
    \midrule
    Command Mean & - & - & - & - & - & 7.98 & 11.41 \\
    UniAD-Single & - & - & - & - & - & 4.16 & 9.31 \\
    BLIP-RT-2 & - & - & - & - & - & 2.78 & 6.47 \\
    \midrule
     & None & $B$ & 35.70 & 43.90 & 65.20 & 2.76 & 6.59 \\
    & $P_1$ & $B$ & 38.20 & 43.67 & 70.74 & 2.67 & 6.41 \\
    \algname & $P_2$ & $B$ & 39.52 & 44.20 & \textbf{78.67} & \textbf{2.62} & 6.19 \\
     & $P_3$ & $B$ & 34.62 & 41.28 & 64.55 & 2.85 & 6.89 \\
     & $P_{1-3}$ & $B$ & \textbf{39.73} & \textbf{54.29} & {70.35} & {2.63} & \textbf{6.17} \\
    \midrule
    \textit{BLIP-RT-2} & \textit{-} & $B$* & \textit{100.0} & \textit{100.0} & \textit{100.0} & \textit{2.41} & \textit{5.79} \\
\bottomrule
\end{tabular}}
\label{tab:waymo_supp}
\vspace{-0.3cm}
\end{table}

\subsection{Computational Complexity}
In Table~\textcolor{red}{5} of the main paper, we provide a comparison of the computational complexity of \algname to UniAD-Single.  
A future direction would be caching the vision tokens and batching the different question patterns, which can speed up the inference time fundamentally.

\section{Qualitative Results}
\label{supp:qualitative}

In this section, we show the qualitative results of the experiments, including VQA on our \dataname-nuScenes, generalization results across sensor data on Waymo, and generalization results to unseen objects on \dataname-CARLA.

\subsection{\dataname-nuScenes}
\begin{figure*}[t!]
    \centering
    \includegraphics[width=\linewidth]{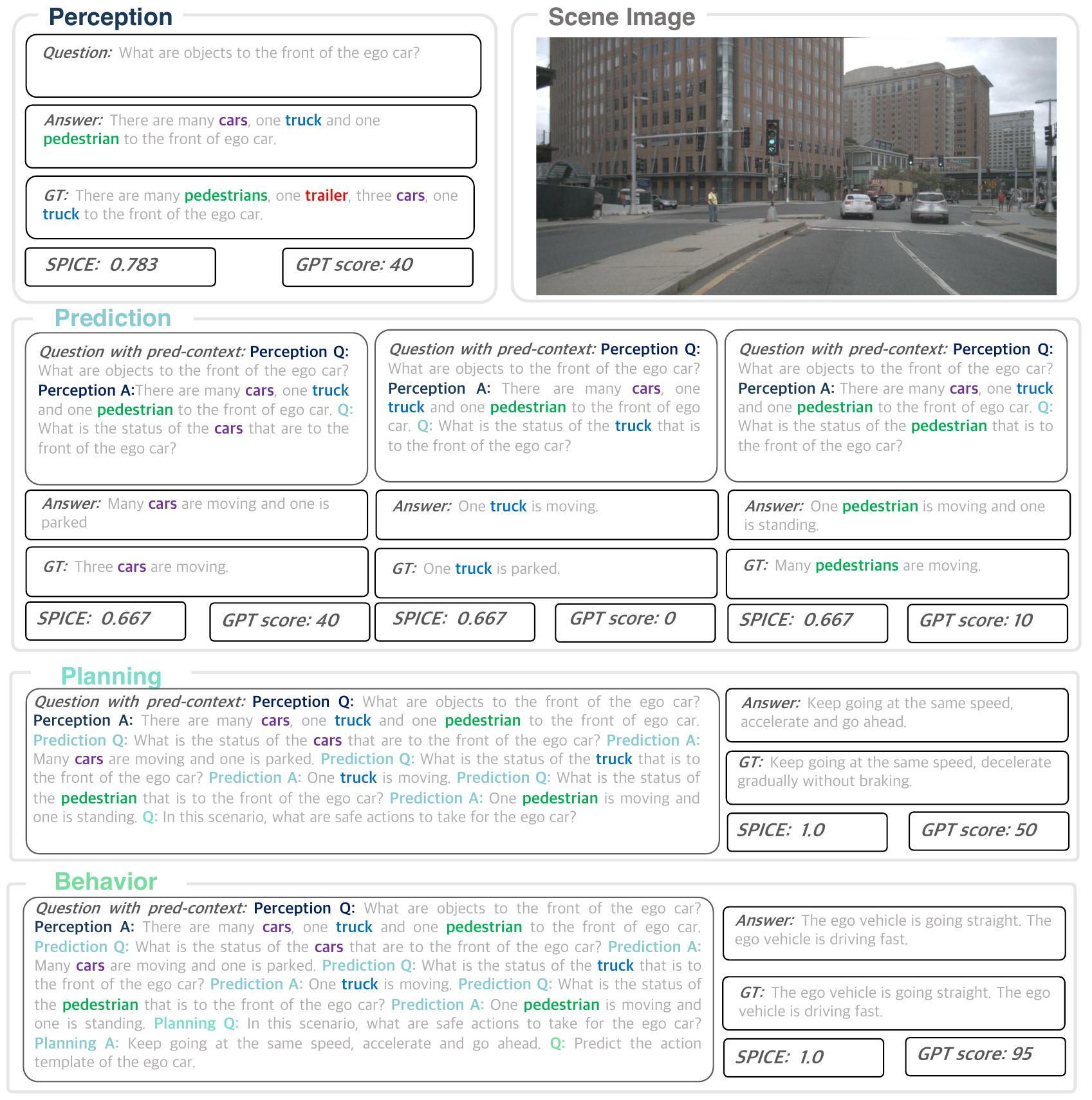}
    \caption{\textbf{Detailed qualitative results on \dataname-nuScenes.} The graph prompting process can be divided into different tasks, and different QAs in each task revolve around different objects.}
    \label{fig:nus_detailed_results}
\end{figure*}

\begin{figure*}[t!]
    \centering
    \includegraphics[width=\linewidth]{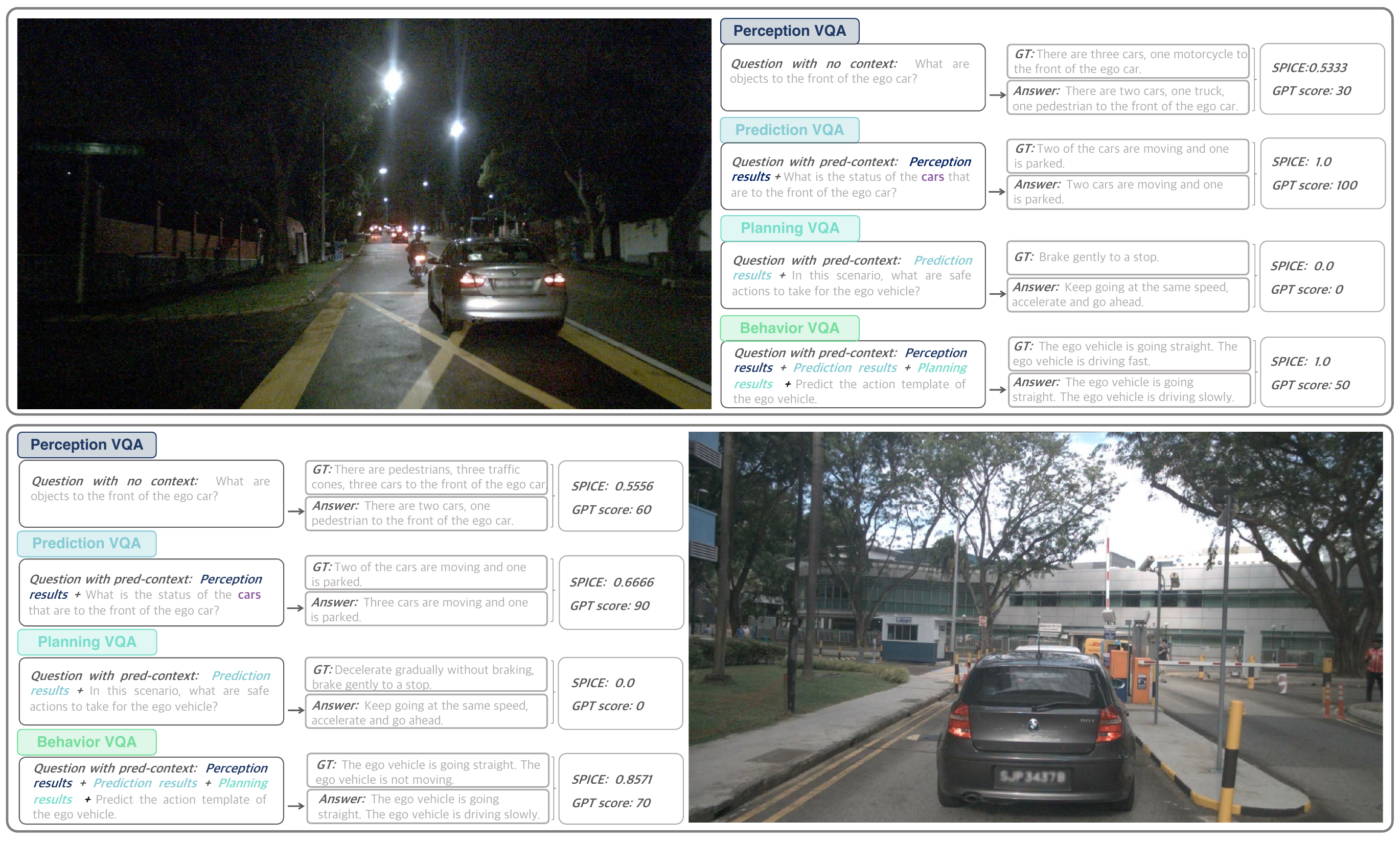}
    \caption{\textbf{More qualitative results on \dataname-nuScenes.} The examples in the figure illustrate the robust ability of our DriveLM-Agent to perform VQA tasks in driving scenarios.}
    \label{fig:nus_results}
\end{figure*}

This section shows the qualitative examples for the \dataname-nuScenes.
In Fig.~\ref{fig:nus_detailed_results}, we showcase a detailed example of GVQA reasoning process on DriveLM-nuScenes, encompassing $P_{1-3}$ QA and the behavior task.
We compare the predicted answers with ground truth and provide SPICE scores and GPT scores.
In this figure, the second question in the prediction stage represents a typical error. 
Due to the input of single-frame images, our model often struggles to accurately determine the correct movement status of objects. 
This judgment is indeed challenging even for humans.
Furthermore, in Fig.~\ref{fig:nus_results}, we present additional qualitative results to showcase our model's performance.


\subsection{Waymo}
\begin{figure*}[t!]
    \centering
    \includegraphics[width=\linewidth]{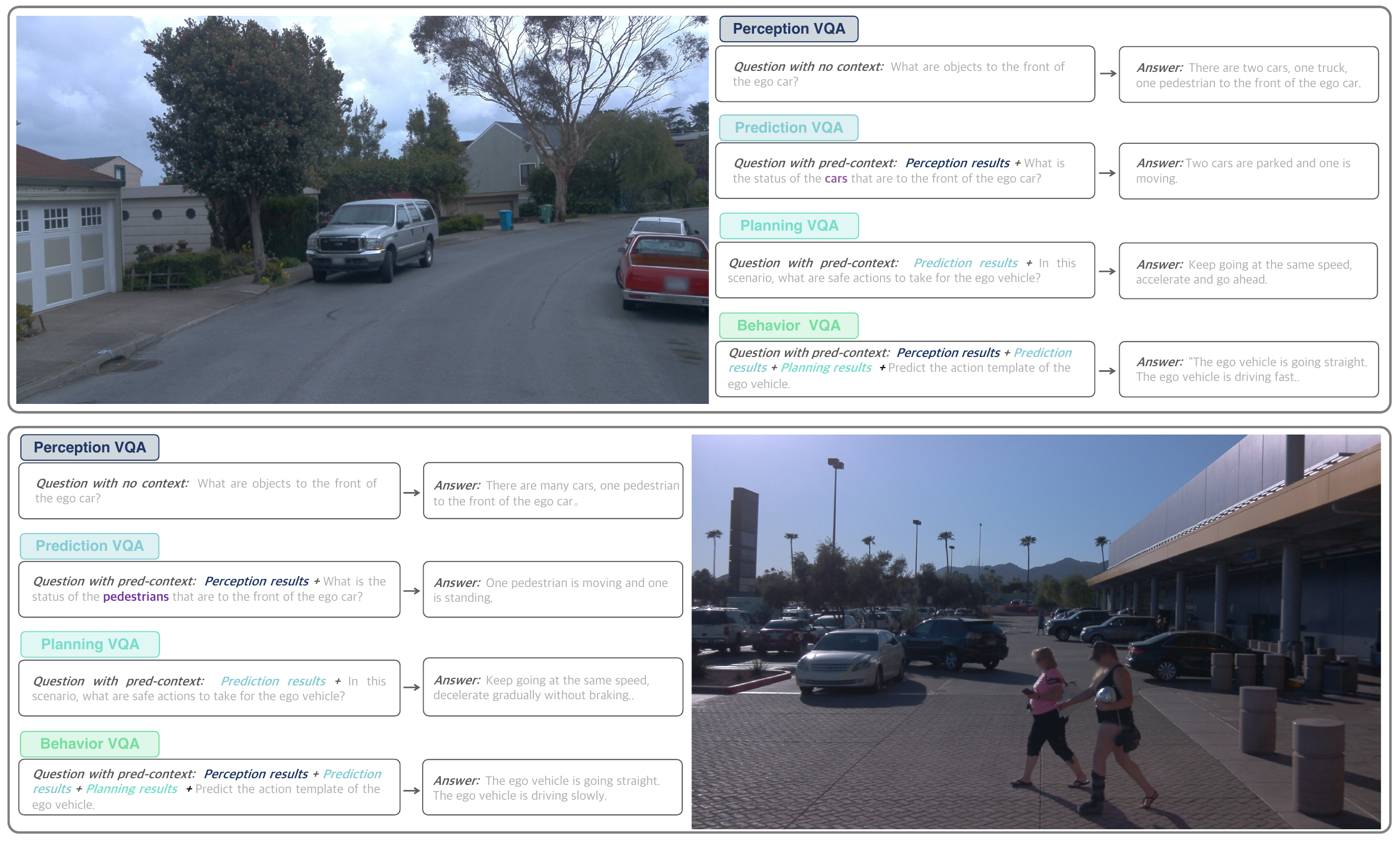}
    \caption{\textbf{Qualitative results on Waymo.} We present two examples showcasing the generability of DriveLM-Agent to new sensor configurations, demonstrating the strong generalization capability of DriveLM-Agent.}
    \label{fig:waymo_results}
\end{figure*}

This section demonstrates the generalizability of our model across sensor configurations. 
Fig.~\ref{fig:waymo_results} illustrates the results of our model, trained on DriveLM-nuScenes, when applied to inference on Waymo. 
As we do not annotate data on Waymo, the questions are manually defined, and ground truth is not provided. 
These results showcase the robust generalization capability of our model.

\subsection{\dataname-CARLA}
In this section, we provide qualitative examples for the CARLA dataset.

\begin{figure*}[t!]
    \centering
    \includegraphics[width=0.9\linewidth]{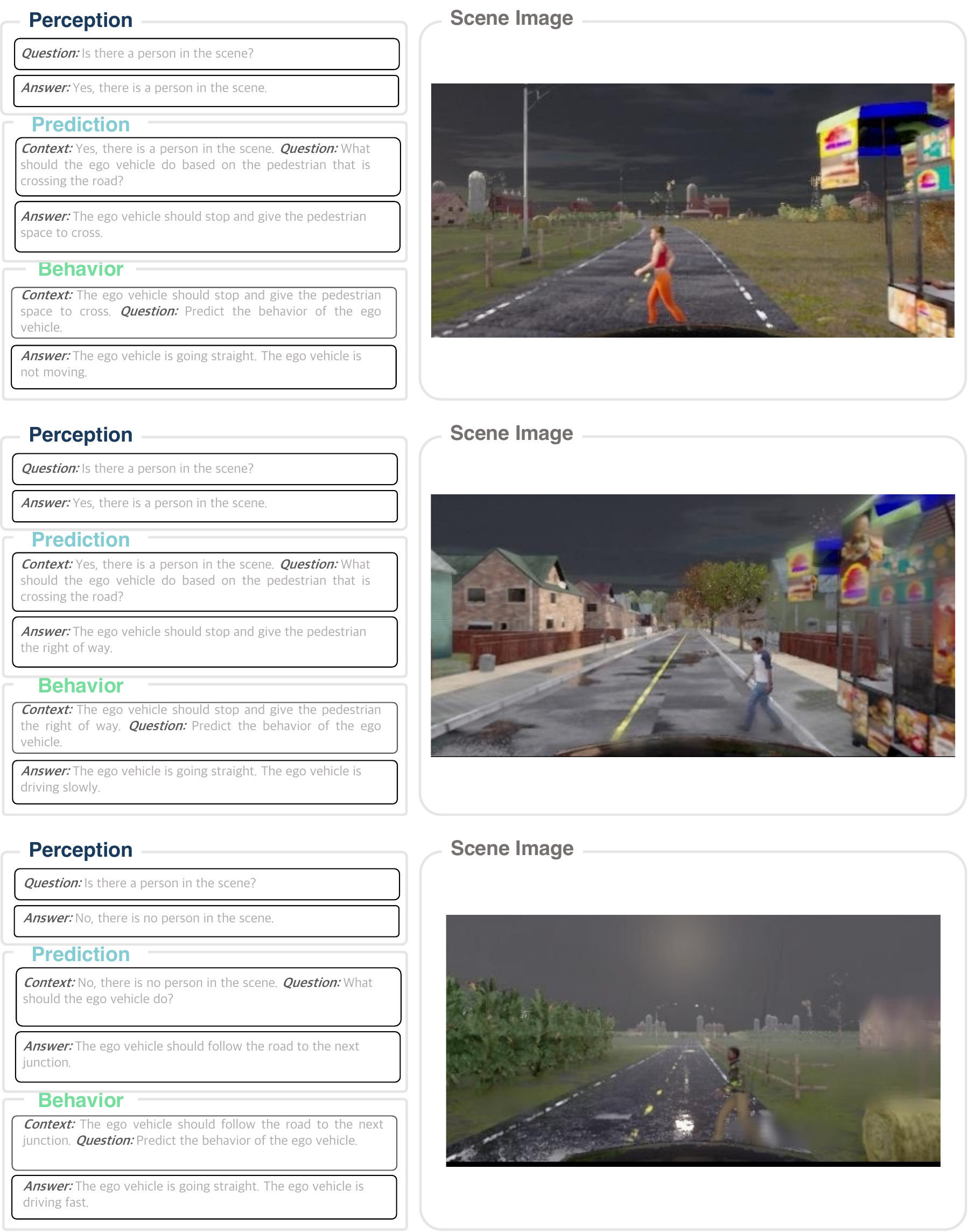}
    \caption{\textbf{Qualitative results on the generalization test set in \dataname-CARLA.} We show three examples of \algname handling the pedestrian scenario. The first example shows a success case and the second and third show two common failure cases of the model.  }
    \label{fig:carla_resultsPed}
\end{figure*}

\smallskip
\bheading{Generalization to the unseen pedestrian scenario.}
Fig.~\ref{fig:carla_resultsPed} shows the generated behaviors for the generalization test set on the unseen pedestrian scenario. 
The first example, illustrated on the top of Fig.~\ref{fig:carla_resultsPed}, demonstrates a successful situation where \algname accurately recognizes a pedestrian. It subsequently infers the appropriate action to undertake, which in this case, is to stop the vehicle. The behavior generation is able to interpret this context, resulting in the correct behavior pattern, as evidenced by the ego vehicle coming to a complete stop.
The other two examples represent the predominant failure modes of \algname in scenarios involving pedestrians. The middle example of Fig.~\ref{fig:carla_resultsPed} shows the case where the model still detects the pedestrian. However, it fails to translate this detection into the correct behavior.
The final example, shown at the bottom, highlights a more severe limitation where the model completely overlooks the pedestrian. In such instances, \algname acts as if the pedestrian is non-existent, which consequently results in it not executing any evasive or stopping maneuvers, posing a significant risk in a real-world scenario. 

\begin{figure*}[t!]
    \centering
    \includegraphics[width=0.9\linewidth]{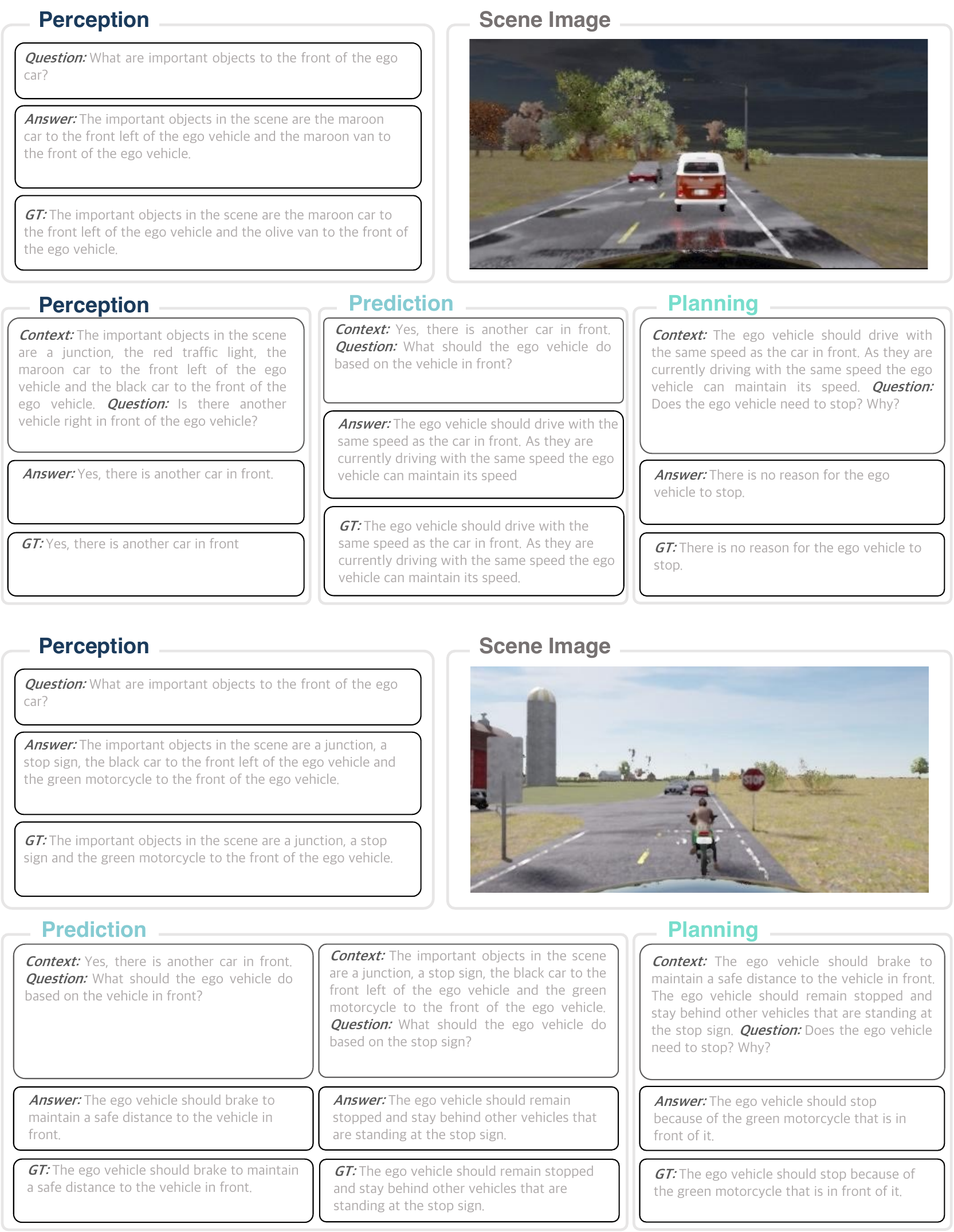}
    \caption{\textbf{Qualitative VQA results in \dataname-CARLA.} The first example shows how \algname deals with a situation with a vehicle directly in front of the ego vehicle. The second example shows an intersection with a stop sign and other traffic participants waiting in front of the ego vehicle at the stop sign.}
    \label{fig:carla_resultsQA}
\end{figure*}

\smallskip
\bheading{Graph Visual Question Answering.}
This section presents two examples of the graph visual question answering tasks using the CARLA dataset to evaluate the performance of \algname (Fig.~\ref{fig:carla_resultsQA}). We only show a subset of the evaluated questions.
In the first example the ego vehicle drives behind another vehicle. The primary task is to follow the road and adjust the speed in accordance with the leading vehicle. Our results indicate that \algname demonstrates proficient scene understanding by accurately identifying all important objects in the scene. Despite the ground truth data indicating occasional inaccuracies in object color labeling by the CARLA simulator, \algname maintains reliable performance in object recognition. 
Additionally, the model can 
identify the vehicle in front and 
reason about what to do based on the leading vehicle.

The second example takes place at an intersection regulated by a stop sign. \algname identifies all objects and can reason about the situation. It correctly identified that it needs to stop not simply due to the stop sign, but primarily because of a motorcycle positioned ahead. This implies that \algname is capable of prioritizing dynamic obstacles over traffic control devices under certain circumstances.

\clearpage

\section{More Related Work}
\label{supp:related}

In this section, we supplement the related work mentioned in the main paper.

\smallskip
\bheading{Reasoning Over Graph Structures.}
Reasoning is one of the basic forms of simulated human thinking, enabling the derivation of new judgments from one or several existing judgements~\cite{chen2020review}. 
Many prior reasoning works have been grounded in graph-based approaches~\cite{shi2019scenegraph,besta2023got,wang2019knowledgegraph,chen2019graph}.
XNMs~\cite{shi2019scenegraph} employs scene graphs for explainable and explicit reasoning with structured knowledge.
KPRN~\cite{wang2019knowledgegraph} utilizes knowledge graph for reasoning and applies it to recommender systems.
GoT~\cite{besta2023got} models LLM-generated information as an arbitrary graph and brings the LLM reasoning closer to brain mechanisms.
Inspired by these successful attempts, we try to link the stages of perception, prediction, and planning in autonomous driving through a graph, 
enabling the model to grasp the reasoning process and deduce unseen scenarios based on learned graph structure.

\bheading{Embodied Planning with LLMs} Recent work endeavors to leverage the formidable reasoning and generalization capacity of LLMs~\cite{touvron2023llama,floridi2020gpt,kojima2022large} for embodied AI systems~\cite{brohan2023can,huang2022inner,zitkovich2023rt,driess2023palme,liang2023code,karamcheti2023voltron,rana2023sayplan}. PaLM-E~\cite{driess2023palme} trains an LLM for various embodied tasks including sequential robotic manipulation planning. CaP~\cite{liang2023code} provides a robot-centric formulation of language model generated programs executed on real systems. RT-2~\cite{zitkovich2023rt} represents robot actions as language tokens, training vision-language models to output robot policies. These methods showcase the capabilities of LLMs in embodied planning tasks, inspiring us to apply them to address the current shortcomings in generalization in AD, which is far less explored.

\smallskip
\bheading{Vision-language Benchmarks for Driving.}
An increasing number of vision-language datasets have been proposed for AD systems~\cite{zhou2023vision,Dewangan2023talk2bev,wu2023referkitti,wu2023nuprompt,qian2023nuscenesqa,kim2019had,kim2018bddx,sachdeva2023rank2tell,malla2023drama,marcu2023lingoqa,ding2024holistic}. 
NuScenes-QA~\cite{qian2023nuscenesqa} and NuPrompt~\cite{wu2023nuprompt} provide perceptual information as text by describing the positions and states of surrounding objects.
BDD-X~\cite{kim2018bddx} provides reasons for the ego vehicle’s actions in natural language descriptions.
DRAMA~\cite{malla2023drama} and Rank2Tell~\cite{sachdeva2023rank2tell} identify crucial objects and provide corresponding driving suggestions.
However, these datasets focus on scene-level context or individual objects.
\dataname fills this gap in the literature by organizing language annotations from object-level and task-level with a graph structure.

\section{Broader Impact}
We believe that this approach can accelerate progress in the field of autonomous driving by enabling it to directly benefit from better VLMs.
Our goal is to make progress towards autonomous driving, which will have profound impact if successful. We recognize that by bringing VLMs into this area, we accept their ethical implications, such as hallucinations and high resource use. Yet, by improving the interactivity between humans and autonomous driving systems, we can build confidence in the technology. This could hasten its acceptance and lead to safer transportation in the long term.

\end{document}